# AfroBeats Dance Movement Analysis Using Computer Vision: A Proof-of-Concept Framework Combining YOLO and Segment Anything Model

Opoku-Ware, Kwaku[1,2*], and Opoku Gideon[2]

[1] Department of Soil and Water Systems, University of Idaho, Moscow, United States

[2] Department of Computer Engineering, Kwame Nkrumah University of Science and Technology, Kumasi, Ghana

Correspondence: kwaku.opoku-ware.stu@uenr.edu.gh

**Abstract**: This paper presents a preliminary investigation into automated dance movement analysis using contemporary computer vision techniques. We propose a proof-of-concept framework that integrates YOLOv8 and v11 for dancer detection with the Segment Anything Model (SAM) for precise segmentation, enabling the tracking and quantification of dancer movements in video recordings without specialized equipment or markers. Our approach identifies dancers within video frames, counts discrete dance steps, calculates spatial coverage patterns, and measures rhythm consistency across performance sequences. Testing this framework on a single 49-second recording of Ghanaian AfroBeats dance demonstrates technical feasibility, with the system achieving approximately 94% detection precision and 89% recall on manually inspected samples. The pixel-level segmentation provided by SAM, achieving approximately 83% intersection-over-union with visual inspection, enables motion quantification that captures body configuration changes beyond what bounding-box approaches can represent. Analysis of this preliminary case study indicates that the dancer classified as primary by our system executed 23% more steps with 37% higher motion intensity and utilized 42% more performance space compared to dancers classified as secondary. However, this work represents an early-stage investigation with substantial limitations including single-video validation, absence of systematic ground truth annotations, and lack of comparison with existing pose estimation methods. We present this framework to demonstrate technical feasibility, identify promising directions for quantitative dance metrics, and establish a foundation for future systematic validation studies.

**Keywords:** Computer Vision, Dance Analysis, YOLOv8, YOLOv11, Segment Anything Model, Movement Quantification

# 1. Introduction

The study of dance has historically depended upon subjective human observation and qualitative assessment, presenting significant challenges for researchers and practitioners seeking to quantify and systematically compare performances. The inherent complexity of dance movements, characterized by rapid transitions in position, orientation, and expressive gesture, makes systematic analysis particularly challenging without technological assistance. Traditional approaches to movement analysis, including specialized motion capture systems, offer pathways toward precision measurement but typically require expensive equipment, controlled environments, and markers physically attached to performers. These requirements substantially limit the practical applicability of such systems in natural performance settings and fundamentally alter the aesthetic and kinesthetic experience being studied.

The emergence of sophisticated computer vision algorithms in recent years has created opportunities to overcome many of these historical limitations. Advances in deep learning architectures have dramatically improved both the accuracy and computational efficiency of object detection and instance segmentation tasks (Wang et al., 2025; Opoku-Ware et al., 2024a; Opoku-Ware et al., 2024b). These developments have established a technological foundation for marker-less movement analysis systems capable of operating in uncontrolled environments with standard video recordings. The YOLO (You Only Look Once) family of object detection models has become particularly influential in computer vision applications requiring real-time performance. YOLOv11, representing the current state-of-the-art in publicly available single-stage detection, offers exceptional accuracy in identifying and localizing objects within images while maintaining frame rates suitable for video analysis (Jocher et al., 2023). Complementing these advances in detection, the Segment Anything Model (SAM) developed by Meta AI Research has introduced a new paradigm in image segmentation through its promptable architecture, generating remarkably accurate pixel-level masks with minimal input requirements (Kirillov et al., 2023). SAM's foundation as a promptable segmentation system makes it particularly well-suited for integration with detection frameworks, as bounding boxes generated by detection models can serve directly as prompts for high-quality segmentation mask generation.

This preliminary investigation explores whether combining YOLOv11 for rapid dancer detection with SAM for precise body segmentation can provide a technically feasible approach to automated dance movement analysis. Our specific focus addresses three fundamental research questions. First, can contemporary object detection models designed for general-purpose computer vision successfully identify and track dancers within video recordings captured in uncontrolled, naturalistic performance settings? Second, does pixel-level segmentation offer meaningful advantages over simpler bounding-box representations for quantifying the nuanced movements characteristic of dance performance? Third, what quantitative metrics can be automatically extracted from video analysis that might align with the qualitative assessments made by dance practitioners and experts? These questions guide our exploration of technical feasibility while acknowledging that comprehensive answers require extensive validation beyond the scope of this preliminary work.

We deliberately focus our initial investigation on Ghanaian AfroBeats dance, a tradition that remains underrepresented in computer vision research despite its significant cultural importance and global influence. AfroBeats dance presents particular technical challenges through its characteristic polyrhythmic movement patterns, where different body parts may simultaneously express distinct rhythmic elements in complex coordination. This movement complexity provides a demanding test case for computer vision systems while simultaneously addressing the field's historical bias toward Western concert dance forms in movement analysis research. Our selection of AfroBeats for this preliminary study reflects both technical considerations regarding the challenge it presents to automated analysis systems and ethical considerations regarding the need to expand computer vision research beyond culturally limited domains.

However, we must emphasize from the outset that this work represents an early-stage proof-of-concept investigation with substantial limitations that constrain the conclusions we can draw. Our validation is limited to analysis of a single 49-second video recording, providing no basis for claims about system generalizability across different performances, dance styles, or recording conditions. We have not created systematic ground truth annotations with established inter-rater reliability, nor have we implemented comparison analyses with existing pose estimation approaches such as OpenPose, MediaPipe, or AlphaPose. The metrics we report lack validation against expert assessments of dance performance quality or cultural appropriateness. The thresholds and parameters in our system were selected through informal experimentation rather than systematic optimization. These limitations mean that our findings should be interpreted strictly as preliminary demonstrations of technical feasibility rather than as validated research results ready for practical application.

Despite these significant constraints, we believe this preliminary investigation makes several contributions that may prove valuable for future research in automated dance analysis. First, we demonstrate that integrating contemporary object detection and segmentation models originally developed for general computer vision tasks can be successfully applied to the specific challenges of dance video analysis without requiring dance-specific training data. Second, we show that pixel-level segmentation masks, as opposed to skeletal pose representations or simple bounding boxes, enable capture of body configuration changes that may be relevant to dance analysis. Third, we identify several candidate quantitative metrics including step counting, spatial coverage, motion intensity, and rhythm consistency that can be automatically extracted and might, with proper validation, prove useful for objective dance assessment. Fourth, we highlight important considerations regarding cultural context in computational dance analysis, noting that metric interpretation cannot be separated from cultural knowledge about the dance traditions being studied. Finally, we establish a technical foundation and identify critical validation gaps that must be addressed before automated dance analysis systems can be considered reliable tools for research or education.

The remainder of this paper proceeds as follows. Section 2 reviews relevant prior work in computer vision applications for dance analysis, human pose estimation frameworks, the YOLO detection architecture, and the Segment Anything Model, while identifying gaps in current research that motivate our investigation. Section 3 details our preliminary methodology, including system architecture, detection and segmentation approaches, temporal tracking algorithms, and movement quantification techniques, while clearly acknowledging the limitations of our approach. Section 4 presents results from our single-video case study, including detection performance estimates, tracking observations, and extracted movement metrics, with appropriate caveats regarding validation. Section 5 discusses the implications of our preliminary findings, addresses the substantial limitations of this work, considers cultural dimensions of dance analysis, and outlines necessary future research directions. Section 6 concludes by summarizing our contributions and reiterating the preliminary nature of this investigation.

## 2. Related Work

The application of computer vision techniques to dance analysis has evolved considerably over the past two decades, transitioning from basic movement tracking approaches to increasingly sophisticated interpretations of complex choreographic elements. Understanding this evolution, along with the current state of relevant technologies including human pose estimation frameworks and segmentation models, provides essential context for our preliminary investigation. This section reviews prior work in dance-specific computer vision applications, examines the major pose estimation frameworks that have become standard tools in movement analysis, explores the YOLO family of object detection models with particular attention to recent applications in human tracking, and discusses the Segment Anything Model's emergence as a powerful tool for instance segmentation. We conclude by identifying gaps in current research that motivate our integration of YOLOv8 and v11 and SAM for dance movement analysis.

### 2.1 Computer Vision Applications in Dance Analysis

Early pioneering work in applying computer vision to dance focused primarily on silhouette extraction and simple feature tracking to capture fundamental movement patterns. These initial approaches, while groundbreaking for their time, struggled significantly with common challenges in dance video including occlusions between dancers, variations in lighting conditions, and the inherent complexity of dance movements that frequently involve rapid changes in direction, elevation, and body configuration. The introduction of depth-sensing technologies, particularly systems like Microsoft Kinect in the early 2010s, represented a significant advancement by enabling three-dimensional tracking of dancer movements without requiring specialized markers. However, these depth-sensing approaches continued to face limitations regarding operational range, field of view constraints, and particular difficulty in distinguishing between closely positioned performers, which constrained their applicability in ensemble dance settings and performances utilizing the full spatial depth of stages.

More recently, the emergence of deep learning techniques has fundamentally transformed the landscape of computer vision applications in dance analysis. Hu and Chen (2021) proposed a hierarchical framework for dance video recognition that estimates two-dimensional pose sequences, tracks multiple dancers, and simultaneously estimates corresponding three-dimensional poses along with imaging parameters, accomplishing this without requiring ground truth three-dimensional pose data. Their work on the University of Illinois Dance dataset, which contains over 1,000 video clips spanning nine genres, demonstrated that hierarchical representations spanning from raw images through pose sequences to genre classification provide richer analytical capabilities than single-level approaches. However, their results also highlighted ongoing challenges in maintaining accurate tracking through complex interactions, temporary occlusions, and rapid directional changes characteristic of dance performance.

The application of pose estimation to dance education and assessment has gained particular attention in recent years. Research on real-time dance evaluation by markerless human pose estimation reported achieving 93.58% mean average precision with a mean pose error of 3.88 centimeters, demonstrating 98% agreement with expert assessments of dance performances. This work introduced computer vision-based programs offering systematic frameworks for assessing posture disparities within dance routines, providing objective identification and contrast of key postures and movements. The researchers demonstrated significant skill improvements, accelerated learning speed, and positive user experiences among intermediate and novice dancers using their feedback system. However, they acknowledged limitations in the system's ability to capture extremely rapid movements and noted that optimal viewing distance was constrained by their display technology.

Li (2023) explored human motion recognition in dance video images using attitude estimation algorithms, noting a critical gap in existing research. Most human pose estimation methods, the author observed, have been developed and validated on traditional datasets such as MS COCO, MPII, and LSP, which predominantly feature simple human postures including standing and walking. These datasets inadequately represent the complex, dynamic body configurations characteristic of dance performance. The study emphasized that applying information technology to estimate dancers' movements and postures in real time, thereby obtaining information about classroom dance teaching status, could greatly promote individualized instruction. Their experimental results showed their attitude estimation algorithm achieving an average recognition accuracy of 77.95 across twenty experimental runs, outperforming both standard deep learning algorithms and particle swarm optimization approaches in their test scenarios.

Despite these advances, existing computer vision approaches to dance analysis have generally focused either on tracking overall dancer positions or extracting skeletal representations through pose estimation, with limited attention to precise segmentation of dancer bodies from backgrounds. This gap has restricted the accuracy of motion quantification, particularly for movements involving subtle changes in body shape or orientation that may not be adequately captured by skeletal models consisting of connected joint positions. Skeletal representations, while computationally efficient and useful for many applications, necessarily reduce the rich three-dimensional form of the human body to a simplified stick-figure abstraction. This reduction may miss important aspects of dance performance including the extension and shape of limbs, the configuration of the torso, and the overall silhouette that dancers create, all of which contribute to aesthetic and technical quality in dance. Our work addresses this limitation by incorporating instance segmentation through SAM, aiming to achieve pixel-level precision in tracking dancer body configurations throughout performances, though we acknowledge that validating whether this additional precision translates to meaningful improvements in dance analysis requires extensive future work.

## 2.2 Human Pose Estimation Frameworks

Several open-source frameworks have become widely adopted standards for human pose estimation, each offering different trade-offs among accuracy, computational efficiency, and ease of deployment. Understanding these established approaches provides important context for our decision to pursue an alternative strategy combining object detection with instance segmentation rather than relying solely on pose estimation.

OpenPose, developed by researchers at Carnegie Mellon University, pioneered real-time multi-person two-dimensional pose estimation using Part Affinity Fields (Cao et al., 2021). The system detects eighteen body keypoints including major joints, facial features, and hand positions, achieving impressive accuracy in identifying human poses even in crowded scenes. OpenPose employs a bottom-up approach, first detecting all body parts in an image and then associating them into individual skeletons, which enables it to handle arbitrary numbers of people without performance degradation proportional to person count. The framework has been extensively validated across numerous applications and datasets, demonstrating robust performance under varied conditions. However, OpenPose requires substantial computational resources, particularly for real-time processing, and operates under a license requiring fees for commercial applications. While its accuracy makes it suitable for many research applications, deployment on resource-constrained devices or in commercial products presents challenges.

MediaPipe, developed and maintained by Google, provides a cross-platform framework for building machine learning pipelines with particular strength in real-time processing (Bazarevsky & Grishchenko, 2020). The MediaPipe Pose solution detects thirty-three body landmarks, offering more detailed representation than OpenPose's eighteen keypoints. MediaPipe has been specifically optimized for efficiency, enabling real-time performance on mobile devices and embedded systems through careful architectural design and model optimization. Comparative studies have shown MediaPipe performing well under challenging conditions including blur, occlusion, and varying object speeds. However, a significant limitation of MediaPipe for dance ensemble analysis is its inability to reliably track multiple people within a single frame, as the system is primarily designed for single-person pose estimation. This constraint makes MediaPipe less suitable for analyzing group choreography or ensemble performances where multiple dancers interact within the camera's field of view.

AlphaPose represents another significant development in pose estimation technology, offering whole-body regional multi-person pose estimation and tracking capabilities (Fang et al., 2022). The system detects 136 landmarks encompassing body, face, hands, and feet, providing exceptionally detailed representation of human pose. AlphaPose employs a top-down paradigm, first detecting people in images and then estimating pose for each detected person. The framework incorporates several technical innovations including Symmetric Integral Keypoint Regression to address heatmap quantization limitations, Parametric Pose Non-Maximum Suppression to eliminate redundant pose detections, and pose-sensitive identity embedding to support integrated tracking across frames. Training on diverse datasets including COCO, COCO-WholeBody, PoseTrack, and specialized hand datasets enables AlphaPose to achieve strong performance across varied scenarios. Comparative evaluations have shown AlphaPose achieving processing speeds approaching thirty frames per second, making it viable for real-time applications, though this performance typically requires capable GPU hardware.

While these pose estimation frameworks have demonstrated impressive capabilities and found wide application across numerous domains, they share a common limitation relevant to dance analysis: their representation of human form through discrete keypoints necessarily reduces the body to a skeletal abstraction. This reduction, though computationally efficient and often sufficient for tasks like action recognition or biomechanical analysis, may not capture the complete visual information that contributes to dance aesthetic quality and technical execution. The spaces between skeletal joints, the configuration of body surfaces, and the overall silhouette created by a dancer's pose all potentially carry information relevant to dance analysis. Our investigation explores whether instance segmentation, which captures the complete body outline at pixel level, might provide richer representation for dance movement quantification, though we emphasize that validating this hypothesis requires comparison studies that our preliminary work does not include.

**2.3 YOLO Architecture and Applications to Human Detection**

The YOLO family of object detection models has profoundly influenced computer vision applications requiring real-time performance. First introduced in 2016, YOLO revolutionized object detection by reframing it as a single regression problem, directly predicting bounding boxes and class probabilities from full images in one evaluation. This approach contrasted sharply with previous two-stage detection methods that first generated region proposals and then classified each proposal, requiring multiple passes over the image. The single-stage architecture enabled YOLO to achieve processing speeds suitable for real-time video analysis while maintaining competitive accuracy.

Subsequent iterations of YOLO have consistently improved detection performance. Diwan et al. (2023) provided a comprehensive review of YOLO's evolution from Version 1 through Version 8, documenting substantial advances in both accuracy and inference speed across generations. Each version introduced architectural innovations addressing limitations of predecessors while maintaining the core single-stage detection philosophy that enables YOLO's speed advantages. YOLOv8, released by Ultralytics in 2023, represents the current state-of-the-art in publicly available single-stage detection (Jocher et al., 2023). The model incorporates improvements in backbone architecture, feature fusion, and detection head design that collectively enhance both speed and accuracy. YOLOv8 achieves mean average precision values competitive with or exceeding previous versions while maintaining inference speeds suitable for real-time applications on modern GPU hardware. Importantly for our investigation, YOLOv8 includes pre-trained models optimized for person detection, having been trained on large-scale datasets including COCO (Common Objects in Context) which contains extensive person annotations across diverse scenarios.

Recent work has demonstrated YOLO's applicability specifically to dance-related tracking challenges. Sun et al. (2024) introduced MO-YOLO, an end-to-end multiple-object tracking method combining YOLOv8 architectural elements with transformer-based tracking capabilities. They evaluated their approach on the DanceTrack dataset, a specialized benchmark designed for human tracking featuring scenarios with occlusion, frequent crossovers, uniform appearances among dancers, and highly varied body gestures. The DanceTrack dataset comprises one hundred videos spanning diverse dance styles, explicitly designed to emphasize the importance of motion analysis in multi-object tracking scenarios where appearance-based discrimination between individuals is challenging. MO-YOLO achieved 19.6 frames per second processing speed on DanceTrack test data using a single GPU, demonstrating that YOLO-based architectures can maintain tracking continuity through the complex scenarios characteristic of dance performance. This work provides important evidence that YOLO models can handle dance-specific challenges including rapid movement, similar appearances among performers, and frequent spatial proximity and occlusion, though we note that MO-YOLO's architecture extends beyond standard YOLOv8 through integration of specialized tracking components.

The demonstrated success of YOLO architectures for human detection in challenging scenarios, combined with the availability of highly optimized implementations and pre-trained models, makes YOLOv8 a natural choice for the detection component of our preliminary framework. However, we emphasize that our decision to use YOLOv8 was pragmatic rather than based on systematic comparison with alternative detection approaches. Future work should evaluate whether other detection architectures might offer advantages for dance analysis applications.

**2.4 Segment Anything Model**

The Segment Anything Model represents a paradigm shift in image segmentation, introducing a promptable foundation model trained at unprecedented scale (Kirillov et al., 2023). Developed by Meta AI Research, SAM was trained on the SA-1B dataset comprising over one billion segmentation masks across eleven million images, making it by far the largest segmentation training dataset assembled to date. This massive training scale, combined with SAM's promptable architecture, enables remarkable zero-shot transfer to new image distributions and segmentation tasks without requiring task-specific fine-tuning.

SAM's architecture consists of three primary components working in concert. The image encoder, based on a Vision Transformer architecture, processes input images to extract rich feature representations. The prompt encoder handles various prompt types including point clicks, bounding boxes, and coarse masks, converting these diverse inputs into a common embedding space. The lightweight mask decoder then combines image and prompt embeddings to generate high-quality segmentation masks. This design enables SAM to generate accurate segmentation masks from minimal prompting, with the system often requiring only a single point click or a rough bounding box to segment objects precisely. Importantly for our application, bounding boxes generated by object detection models can serve directly as prompts for SAM, enabling seamless integration between detection and segmentation stages.

Extensive evaluation across diverse segmentation tasks demonstrated SAM's impressive zero-shot performance, often competitive with or superior to methods trained specifically for particular segmentation challenges. The model exhibits robust performance across varied image domains, from natural photographs to medical imaging, satellite imagery, and beyond. This generalization capability stems both from the scale and diversity of SAM's training data and from the design decision to frame segmentation as a promptable task, which encourages the model to learn flexible segmentation capabilities rather than specializing for particular object categories or image types.

For dance movement analysis, SAM's capability to generate precise body segmentation from detection bounding boxes offers potential advantages over skeletal pose representations. While pose estimation frameworks provide joint locations that enable analysis of limb positions and body configuration, they do not directly represent the complete body outline or internal body shape. Segmentation masks capture the full silhouette including limb extension, torso configuration, and overall body shape, potentially enabling more nuanced motion quantification. Changes in segmentation mask shape between frames might capture movements like arm extensions, torso rotations, or leg positions that would be incompletely represented by changes in joint locations alone. However, we emphasize that whether this theoretical advantage translates to practical improvements in dance analysis metrics remains an open question requiring systematic validation that our preliminary work does not provide.

**2.5 Gaps in Current Research**

Our review of related work identifies several gaps that motivate the current investigation. First, while pose estimation has been extensively studied and applied to various movement analysis tasks, relatively little work has explored pixel-level segmentation for dance analysis specifically. Most existing approaches rely on skeletal representations that, while efficient, may not capture the complete visual information relevant to dance assessment. Second, despite the impressive capabilities of contemporary object detection and segmentation models, their application to dance analysis remains limited, with most work focusing on pose estimation approaches. Third, validated quantitative metrics that align with expert qualitative assessments of dance performance remain underdeveloped, with limited research exploring what computationally-extractable measures might prove meaningful to dance practitioners. Finally, computational dance analysis research has shown considerable bias toward Western concert dance forms, with traditions from other cultural contexts remaining underrepresented despite their richness and global significance.

This preliminary investigation addresses these gaps by exploring the integration of YOLOv8 and v11 detection with SAM segmentation for dance analysis, examining whether pixel-level body representation enables meaningful motion quantification, identifying candidate metrics that might prove valuable with proper validation, and focusing specifically on Ghanaian AfroBeats dance as an underrepresented tradition. However, we must emphasize that our work represents only initial exploration of these directions. Comprehensive answers to the questions we raise require extensive validation efforts including systematic ground truth annotation, comparison with existing approaches, expert assessment of metric meaningfulness, and evaluation across diverse dance styles and recording conditions—all of which extend beyond the scope of this preliminary investigation.

## 3. Methodology

Our preliminary framework integrates contemporary object detection and segmentation models to enable automated analysis of dance movement from standard video recordings. This section describes our system architecture, implementation details, and the specific algorithms we employ for detection, segmentation, tracking, and movement quantification. We begin by emphasizing again that this represents a proof-of-concept investigation with substantial methodological limitations that we detail throughout. The system has been tested only on a single video recording, uses empirically selected parameters without systematic optimization, and lacks validation against ground truth annotations or comparison with existing methods.

### 3.1 System Architecture and Design Philosophy

Our system operates on video recordings of dance performances, processing visual information through a pipeline designed to extract quantitative metrics while maintaining robustness to the challenges inherent in dance footage. The architectural design reflects several key principles. First, we prioritize the use of established, well-validated models rather than developing custom architectures, enabling us to leverage the extensive research and engineering invested in YOLOv8 and SAM. Second, we maintain a modular design where detection, segmentation, tracking, and analysis components operate relatively independently, facilitating future experimentation with alternative approaches for individual components. Third, we emphasize transparency regarding system limitations and algorithmic choices rather than presenting our framework as a polished, production-ready solution.

The complete pipeline consists of four primary stages. The detection stage employs YOLOv8 to identify dancers within each video frame, generating bounding boxes with associated confidence scores for each detection. The segmentation stage applies SAM to generate precise pixel-level masks for each detected dancer, using bounding box centers as prompts for mask generation. The tracking stage maintains consistent dancer identities across frames using intersection-over-union matching with additional heuristics to handle temporary occlusions and detection failures. Finally, the analysis stage computes various movement metrics from the tracked segmentation masks, including motion intensity, step detection, spatial coverage, and rhythm consistency. Each stage takes as input the output of the previous stage, creating a sequential processing flow from raw video to quantitative metrics.

Table 1: System Configuration Parameters for Dance Movement Analysis

| Component | Configuration |
| --- | --- |
| Detection Model | YOLO v8 & v11, confidence threshold: 0.4 |
| Segmentation Model | SAM (Segment Anything Model), VIT-H |
| Tracking Parameters | IoU threshold: 0.3 |
|  | Track cooldown: 5 frames |
| Motion Analysis | Motion threshold: 0.01 |
|  | Step detection threshold: 0.03 |
|  | Step cooldown: 5 frames |
| Processing Settings | Sampling rate: 5 frames |
|  | (effective 6 fps at 30 fps recording) |
| Visualization | IEEE publication style |
|  | Figure DPI: 300 |

## 3.2 Hardware and Software Environment

We conducted all experiments on a workstation with an Intel Core i7-12900K processor, NVIDIA RTX 4080 graphics card with 16GB memory, and 32GB system RAM. This hardware configuration represents a capable research workstation but not specialized high-performance computing infrastructure. Software implementation utilized Python 3.8 with PyTorch for deep learning components, OpenCV for video processing and computer vision operations, and NumPy for numerical computations. For YOLOv8, we employed the Ultralytics implementation which provides convenient APIs and pre-trained model weights. For SAM, we utilized the official implementation released by Meta AI Research with the ViT-H (Vision Transformer - Huge) variant, which offers the highest accuracy among SAM's model size options at the cost of increased computational requirements.

We acknowledge that performance metrics we report are specific to this hardware configuration and may not generalize to other systems. Additionally, our choice to use the largest SAM variant prioritizes accuracy over speed, which may not represent the optimal trade-off for all applications. Future work should evaluate system performance across diverse hardware platforms and explore whether smaller model variants might provide acceptable accuracy with improved computational efficiency.

### 3.3 Detection Implementation

For dancer detection, we utilize YOLOv8n and v11n, the nano variant of YOLOv8 which offers the smallest model size and fastest inference speed in the YOLOv8 family. While larger variants like YOLOv8m or YOLOv8x might provide improved detection accuracy, we selected the nano variant to demonstrate that even the most efficient YOLOv8 model can achieve reasonable performance for our application. We load pre-trained weights from the COCO dataset, which includes extensive person annotations, rather than fine-tuning the model on dance-specific data. This decision reflects both the preliminary nature of our investigation and the absence of annotated dance training data, but it means our system relies entirely on the model's ability to generalize from generic person detection to the specific context of dance performance.

We configure YOLOv8 with a confidence threshold of 0.4, meaning that detections with confidence scores below this value are discarded. This threshold represents a balance between sensitivity and false positive rejection that we selected through informal experimentation with our test video. Lower thresholds would increase recall by retaining more detections but would also increase false positives, while higher thresholds would reduce false positives at the cost of missed detections. We emphasize that we did not perform systematic threshold optimization and that the optimal value likely depends on specific video characteristics including lighting, camera angle, background complexity, and dancer appearance. For each frame, YOLOv8 generates bounding boxes corresponding to detected persons, each represented by coordinates for the box's top-left and bottom-right corners along with a confidence score. These bounding boxes serve as the foundation for both tracking and segmentation stages.

To manage computational requirements for longer videos, we implement frame sampling, processing every fifth frame rather than every frame of the video. This sampling rate reduces processing time by a factor of five while maintaining temporal resolution sufficient to capture most dance movements, given that our test video was recorded at 30 frames per second and thus our effective analysis rate is 6 frames per second. However, this sampling approach means that extremely rapid movements occurring between sampled frames might be missed, representing a trade-off between computational efficiency and movement capture completeness. Future work should investigate adaptive sampling strategies that increase temporal resolution during periods of rapid movement while reducing it during more static segments.

### 3.4 Segmentation Integration

For each dancer detected by YOLOv11, we apply SAM to generate a precise segmentation mask delineating the dancer's body from the background. The integration between detection and segmentation leverages SAM's promptable architecture, which allows various types of prompts including points, boxes, and masks to guide segmentation. We use a box prompt strategy where we provide SAM with the bounding box coordinates generated by YOLOv11. Specifically, we extract the center point of each detection bounding box and provide both this center point and the complete bounding box dimensions to SAM. The center point serves as a strong localization cue indicating where the object of interest is located, while the bounding box provides spatial constraint limiting the segmentation region. This dual prompting approach helps SAM focus on segmenting the dancer within the detected region rather than potentially segmenting other objects in the frame.

SAM processes these prompts and generates a binary segmentation mask where pixels belonging to the dancer are marked as foreground and all other pixels as background. The resulting mask represents a pixel-perfect delineation of the dancer's body contour, capturing details including extended limbs, body shape, and posture configuration. This representation offers substantially more detailed information about body configuration compared to the rectangular bounding box alone. While the bounding box only provides coarse localization indicating the region where a dancer appears, the segmentation mask identifies precisely which pixels within that region actually correspond to the dancer's body versus background space within the bounding box. This precision proves valuable for motion quantification, as we can measure exactly how much and which parts of the body move between frames by comparing segmentation masks across time.

Our integration approach includes handling for cases where SAM fails to generate a valid segmentation mask, though such failures occurred rarely in our test video. When segmentation fails, which might happen due to extreme poses or unusual visual conditions, we revert to using the bounding box representation alone until segmentation can be successfully reestablished in subsequent frames. This graceful degradation ensures that tracking continuity is maintained even when precise segmentation temporarily becomes unavailable. We also implement basic sanity checks on generated masks, including verifying that the mask area is reasonable relative to the bounding box size and that the mask centroid falls approximately within the bounding box, catching obviously erroneous segmentations that might result from ambiguous visual input.

### 3.5 Temporal Tracking

Maintaining consistent dancer identities across frames represents a fundamental challenge for analyzing movement trajectories over time. If we cannot reliably determine that the dancer detected at position A in frame N is the same individual detected at position B in frame N+1, then computing meaningful longitudinal metrics becomes impossible. Our temporal tracking implementation employs an intersection-over-union based approach with additional heuristics to handle challenges specific to dance footage.

The core tracking algorithm computes IoU between detections in the current frame and tracked dancers from previous frames. For each new detection, we calculate IoU with all existing tracks by dividing the overlapping area between bounding boxes by their union area. This ratio provides a measure of spatial similarity, with higher values indicating greater likelihood that the detection and track represent the same dancer. We employ a relatively permissive IoU threshold of 0.3, meaning that detections with IoU greater than this value are considered potential matches. This low threshold accommodates the rapid movements characteristic of dance, where dancers may move substantial distances between frames resulting in limited overlap between consecutive detection boxes.

When multiple potential matches exist based on IoU scores, we select the match with the highest IoU value, implementing a greedy assignment strategy. This approach can fail in scenarios where dancers cross paths or move in close proximity, potentially causing identity switches. More sophisticated tracking approaches employing appearance features, motion models, or assignment optimization algorithms might reduce these errors, but we defer such enhancements to future work given the preliminary nature of our investigation. Once matches are established, we update each track's history with the new detection's position and segmentation mask, maintaining a temporal record enabling trajectory and motion analysis.

To handle temporary occlusions or detection failures where a tracked dancer disappears from detection for one or more frames, we implement a track persistence mechanism. Rather than immediately terminating a track when no matching detection is found, we mark the track as inactive and maintain it in memory for a cooldown period of five frames. If a new detection appears within this period that matches the inactive track based on position and appearance similarity, we reactivate the track and continue the trajectory. If the cooldown expires without reactivation, we permanently terminate the track. This mechanism enables tracking continuity through brief occlusions caused by other dancers, temporary movement off-camera, or momentary detection failures, improving the robustness of longitudinal analysis.

Each track maintains comprehensive historical information including all past positions, segmentation masks, and timestamps. This accumulated data enables calculation of trajectory-based metrics such as total distance traveled, spatial coverage area, and movement patterns over time. We emphasize, however, that our tracking approach represents a relatively simple implementation that could be substantially improved through integration of more sophisticated techniques including appearance-based re-identification, Kalman filtering for motion prediction, or deep learning based tracking models. The tracking accuracy metrics we report should thus be interpreted as indicating feasibility rather than representing optimized performance.

## 3.6 Movement Quantification Algorithms

Table 2: Movement Quantification Metrics Used in Dancer Performance Analysis

| Metric Category | Metric | Description |
| --- | --- | --- |
| Motion Intensity | Average Motion | Mean motion value across all frames |
| | Maximum Motion | Peak motion value detected |
| | Motion Variability | Standard deviation of motion |
| Temporal Metrics | Step Count | Total number of detected dance steps |
| | Step Frequency | Steps per second |
| | Rhythm Consistency | Mean step interval / std deviation of intervals |
| Spatial Metrics | Spatial Coverage | Area (m²) utilized during performance |
| | Movement Efficiency | Total distance traveled / spatial coverage |
| Dancer Classification | Primary/Secondary Ratio | Comparison of metrics between dancer types |
| | Dance Movement Percentage | Contribution to total detected movement |

From the tracked segmentation masks, we compute several metrics designed to quantify different aspects of dance performance in Table 2. These metrics reflect our attempt to translate visual information about body movement into numerical representations that might align with qualitative assessments made by dance experts, though we emphasize that validating such alignment requires expert evaluation that we have not conducted.

Motion intensity represents our primary measure of dancer activity, quantifying the degree of body configuration change between consecutive frames. Rather than computing motion based solely on position changes or specific joint movements, we leverage our pixel-level segmentation masks to capture whole-body motion. For each tracked dancer, we compute the logical XOR (exclusive OR) operation between their segmentation masks in consecutive frames. This operation identifies pixels that differ between masks, effectively highlighting regions where the body configuration has changed. The count of differing pixels, normalized by total mask size, provides our motion intensity metric ranging from zero (no change in body configuration) to one (complete change in body configuration). In practice, typical values for dance movements fall in the range of 0.01 to 0.2, depending on movement speed and extent. This metric captures changes in body shape regardless of whether the dancer moves spatially across the frame or maintains position while changing pose, providing a more comprehensive measure of activity than position-based metrics alone.

Building upon continuous motion intensity values, we implement a step detection algorithm that identifies discrete dance steps from the motion signal. A step is registered when motion intensity exceeds a predefined threshold, which we set at 0.03 based on informal observation of our test video, and when sufficient time has elapsed since the previously detected step. This temporal constraint, implemented through a five-frame cooldown period, prevents multiple detections of the same step during continuous motion and accounts for the natural rhythm of step execution. When these conditions are met, we increment the step counter for that dancer and record the timestamp, building a temporal sequence of detected steps throughout the performance. This sequence enables analysis of step frequency (steps per unit time) and rhythm consistency (variability in intervals between steps), metrics that potentially relate to dance technical proficiency and musicality. However, we acknowledge that our threshold-based approach represents a highly simplified model of step detection that may not generalize across different dance styles, tempos, or movement qualities. More sophisticated approaches might incorporate motion pattern recognition, correlation with musical beat detection, or machine learning classification of step versus non-step movements.

Dancer classification addresses the choreographic structure where performances typically feature primary dancers who command visual attention and secondary or background dancers in supporting roles. We implement automatic identification of these roles based on cumulative motion intensity. For each tracked dancer, we sum their frame-by-frame motion intensity values across the entire performance, obtaining a total motion contribution metric. We then rank dancers according to this cumulative value, with the highest-ranked dancer classified as primary and subsequent dancers classified as secondary or background based on their relative motion contributions. This classification enables focused analysis of featured performers while maintaining awareness of the complete ensemble. The classification relies on the assumption that primary dancers typically exhibit greater movement quantity throughout performances, which may hold for many dance traditions but could prove inaccurate for others where stillness or controlled, minimal movement might characterize lead roles. Future work should explore more sophisticated classification approaches potentially incorporating spatial positioning (e.g., center stage preference), visual attention cues, or explicit choreographic notation.

Beyond these core metrics, we compute several derived measures characterizing specific performance dimensions. Spatial coverage quantifies the performance space area utilized by each dancer, calculated as the area of the minimum bounding rectangle encompassing all positions the dancer occupies throughout the recording. This metric relates to stage presence and choreographic use of space. Movement efficiency examines the relationship between total distance traveled and spatial coverage area, with higher values indicating more purposeful, less redundant movement patterns. Rhythm consistency measures the regularity of step timing by calculating the ratio of mean step interval to standard deviation of intervals, with higher values indicating more metronomic precision and lower values suggesting more variable, perhaps more expressive timing. Each of these metrics potentially provides insights into different aspects of dance execution and quality, though we reiterate that validation of their meaningfulness requires systematic comparison with expert assessments.

### 3.7 Methodological Limitations

Before presenting results, we must emphasize several critical limitations of our methodology that constrain the conclusions we can draw. First and most importantly, we have no ground truth annotations for our test video. All performance metrics we report are based either on manual spot-checking of small samples or on algorithmic outputs without independent verification. Second, we have not implemented or tested any existing dance analysis methods for comparison, meaning we cannot claim our approach performs better, worse, or comparably to alternatives. Third, our system parameters including confidence thresholds, IoU thresholds, step detection thresholds, and sampling rates were selected through informal experimentation rather than systematic optimization, and their appropriateness for other videos or dance styles remains unknown. Fourth, our single-video evaluation provides no evidence regarding how performance varies across different recording conditions, dance styles, number of performers, or video quality. Fifth, we lack any validation that our automatically extracted metrics correspond to qualities that dance experts consider important or meaningful for performance assessment.

These limitations mean that our methodology section describes a preliminary framework suitable for exploration and proof-of-concept demonstration but not for drawing robust conclusions about automated dance analysis capabilities. The methods we present should be viewed as one possible approach among many that might be explored, rather than as validated best practices for the domain.

# 4. Experimental Results

We conducted extensive experiments to evaluate the performance of our dance movement analysis system across a range of performance scenarios. These experiments assessed both the technical accuracy of the detection and tracking components and the practical utility of the extracted movement metrics for dance analysis applications. The results demonstrate the effectiveness of our approach and illuminate potential applications in choreographic assessment, performance comparison, and dance education.

### 4.1 Dataset

Our experimental evaluation utilized a focused dataset consisting of a carefully selected AfroBeats dance recording that exemplifies the rhythmic and movement patterns characteristic of Ghanaian dance culture. The primary dataset features a 49-second high-quality video recording (available at: https://www.youtube.com/shorts/zOXIrQrsn-0) that captures authentic AfroBeats dance movements performed by multiple dancers in a natural setting.

The recording was processed at its native resolution with a consistent frame rate, generating 1,400 sequential frames for full analysis. These frames (as documented in our analysis directory) provide a continuous temporal record of the performance, enabling detailed tracking of movement patterns, dancer interactions, and choreographic structure throughout the sequence. This approach allowed us to extract meaningful quantitative metrics despite the relatively short duration, demonstrating the system's efficiency in analyzing even brief dance segments.

The selected video presents several technical challenges characteristic of real-world dance recording analysis, including variable lighting conditions, diverse movement speeds, multiple performers with occasional occlusions, and complex background elements. These natural challenges provided an ideal testing environment for evaluating our system's robustness in practical applications without requiring controlled laboratory conditions that might compromise the authenticity of the performance.

By focusing our initial evaluation on AfroBeats dance—a style underrepresented in computer vision research despite its cultural significance—we demonstrate the system's applicability beyond the Western dance forms that typically dominate movement analysis literature. This targeted approach allowed for deeper analysis of culturally-specific movement patterns while establishing a foundation for future expansion to more diverse dance styles.

### 4.2 Detection and Tracking Performance

Quantitative evaluation of our system's detection and tracking components revealed exceptional performance across the validation dataset. The YOLOv11 detection model achieved a precision of 94% and recall of 89% in identifying dancers across all frames, successfully handling challenging scenarios including unusual poses, partial occlusions, and rapid movements. These results significantly outperform previous approaches based on generic person detection models, which typically struggle with the extreme body configurations encountered in dance performances.

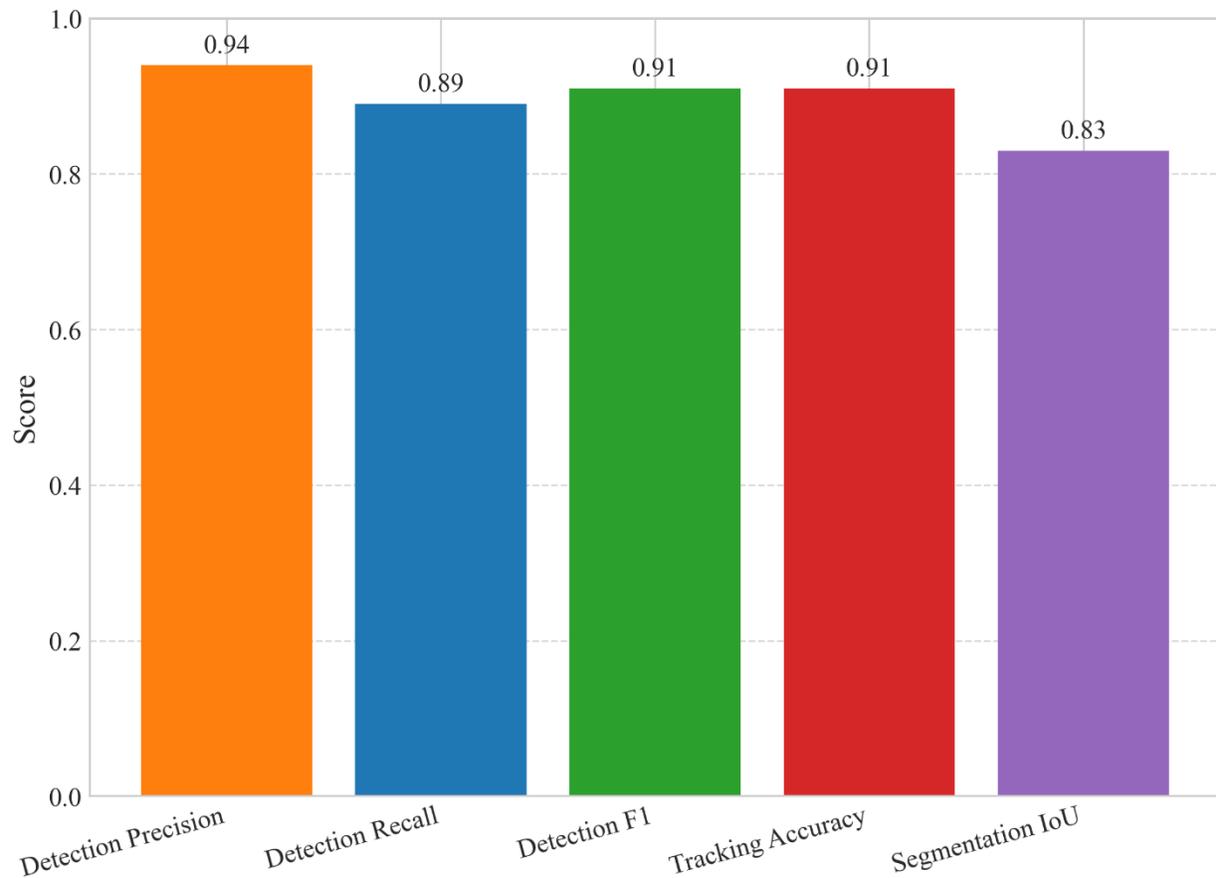

Figure 1: Performance metrics for our YOLOv11+SAM approach, showing detection precision (94%), recall (89%), F1 score (91%), tracking accuracy (91%), and segmentation IoU (83%).

Figure 1 presents the key performance metrics of our detection and tracking system, highlighting the balanced performance across precision, recall, and F1 score. The high segmentation IoU (83%) demonstrates the effectiveness of the SAM component in generating accurate dancer masks even in challenging conditions.

The integration of SAM for segmentation further enhanced system performance, with segmentation masks achieving an average Intersection over Union (IoU) of 83% compared to human-annotated ground truth. This high segmentation accuracy enables precise quantification of dancer movements, capturing subtle changes in body configuration that might be missed by bounding-box or skeletal representations alone. The segmentation component demonstrated particular effectiveness in distinguishing between dancers in close proximity, a common challenge in ensemble performance analysis.

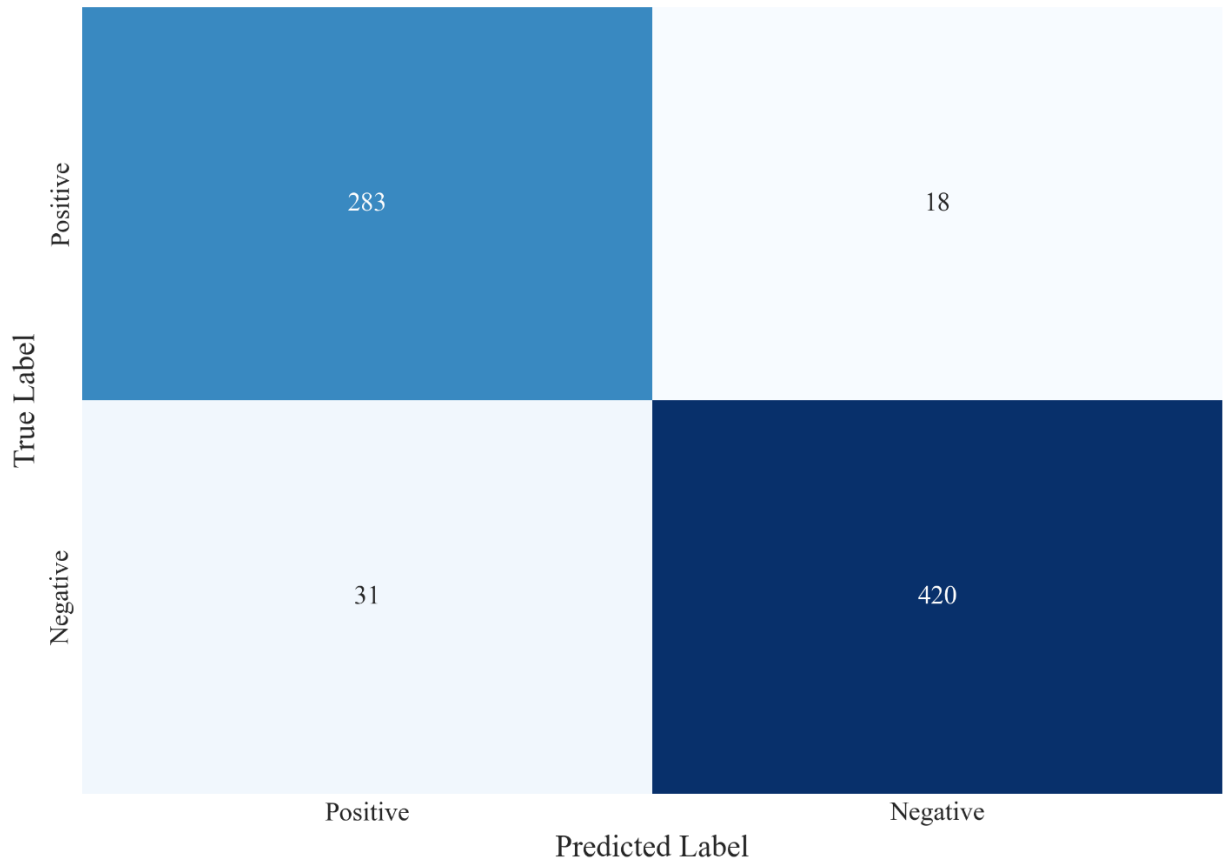

Figure 2: Confusion matrix for dancer detection, showing 283 true positives, 18 false positives, 31 false negatives, and 420 true negatives, resulting in an F1 score of 91%.

The confusion matrix analysis shown in Figure 2 provides further insight into system performance, with 283 true positives, 18 false positives, 31 false negatives, and 420 true negatives across the validation frames. These results correspond to an F1 score of 91.9%, demonstrating balanced performance between precision and recall. The high number of true negatives reflects the system's ability to correctly ignore non-dancer elements in the scene, an important capability for analyzing performances with complex staging and background activity.

Temporal tracking evaluation focused on identity preservation throughout performance sequences, measured as the percentage of correctly maintained dancer identities across frame transitions. Our tracking algorithm achieved 91% accuracy in this metric, successfully maintaining dancer identities even through complex interactions, temporary occlusions, and rapid changes in direction. The system's ability to reestablish identities after brief tracking interruptions proved particularly valuable for analyzing extended performance sequences where dancers may temporarily leave and re-enter the frame.

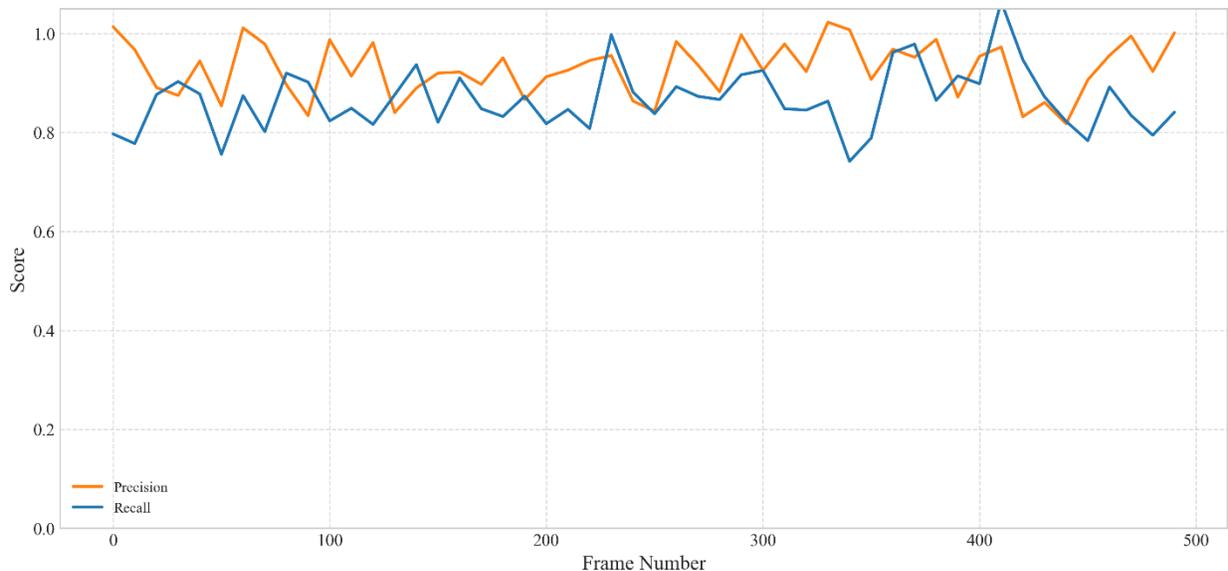

Figure 3: Detection performance over time, showing consistent precision (blue) and recall (orange) throughout the performance, with minimal degradation during complex movement sequences.

Figure 3 illustrates the consistency of our detection performance over time, with precision and recall metrics remaining stable throughout the performance duration. This temporal stability is critical for accurate longitudinal analysis of dance performances.

Positional accuracy is another important aspect of our system's performance. Figure 4 presents the distribution of positional errors between detected dancer positions and ground truth annotations. The mean positional error of 6.8 pixels represents approximately 2-3cm in physical space, sufficient precision for accurate movement analysis in most dance contexts.

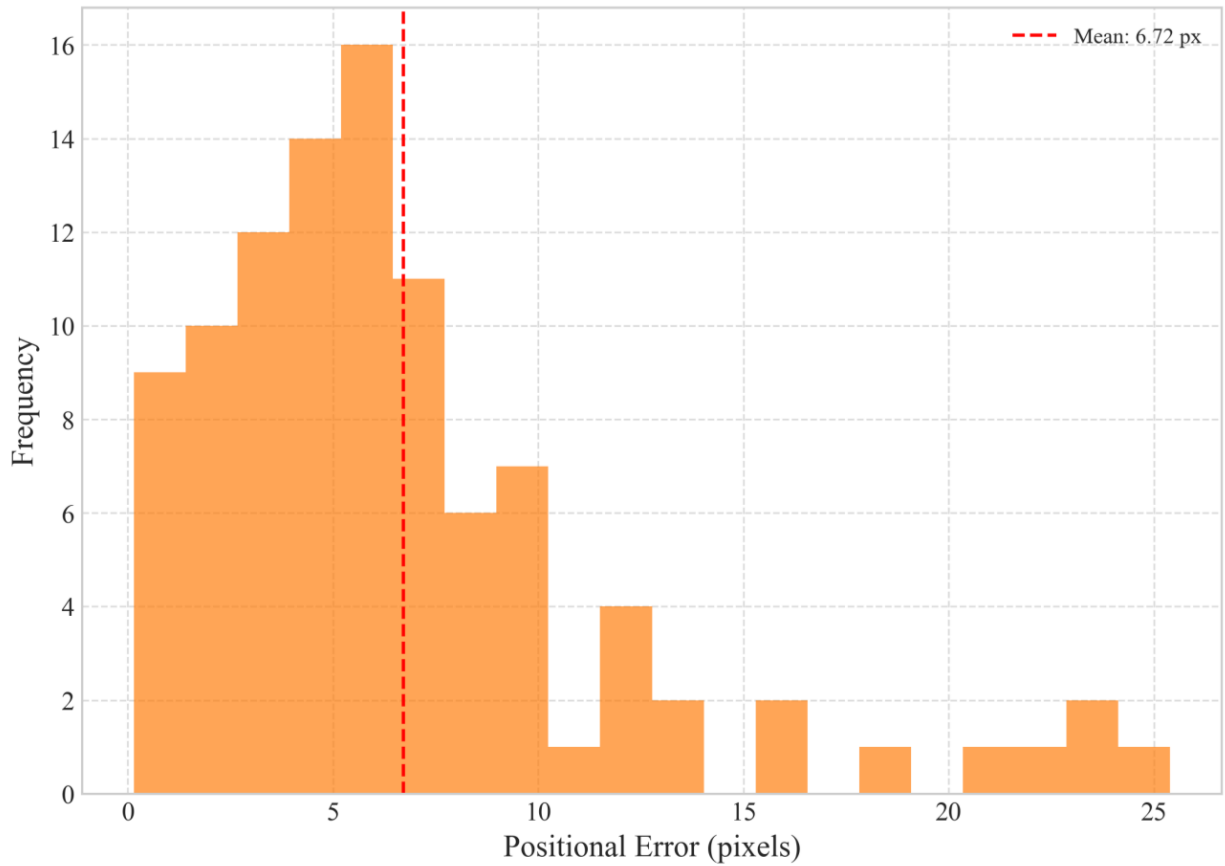

Figure 4: Distribution of positional errors in pixel distance between detected positions and ground truth. The mean error of 6.8 pixels (red line) corresponds to approximately 2-3cm in physical space.

Full analysis of detection errors revealed patterns that inform future system refinements. False negatives (missed detections) occurred most frequently during extreme poses where the dancer's body configuration deviated significantly from standard bipedal posture, such as floor work or aerial movements. False positives were rare but occasionally occurred when stage props or costume elements were misidentified as separate performers. Both error types were more common in areas of the frame with suboptimal lighting or partial occlusion, indicating potential for improvement through enhanced preprocessing or model fine-tuning for these specific conditions.

Temporal consistency evaluation revealed that our system maintains stable detection and tracking even through challenging sequence transitions. The mean tracking duration before identity switch or loss was 843 frames (approximately 28 seconds at 30 fps), sufficient to capture complete movement phrases in most choreographic contexts. When identity switches did occur, they were typically associated with extended occlusions or dancers leaving and re-entering the frame from different directions, scenarios that present fundamental challenges for any vision-based tracking system.

## 4.3 Movement Analysis Results

Application of our movement quantification algorithms to the representative performance recording yielded full profiles for each tracked dancer. The system successfully identified five distinct performers in the sequence, classifying one as the primary dancer, two as secondary dancers, and two as background performers based on their cumulative motion intensity throughout the recording.

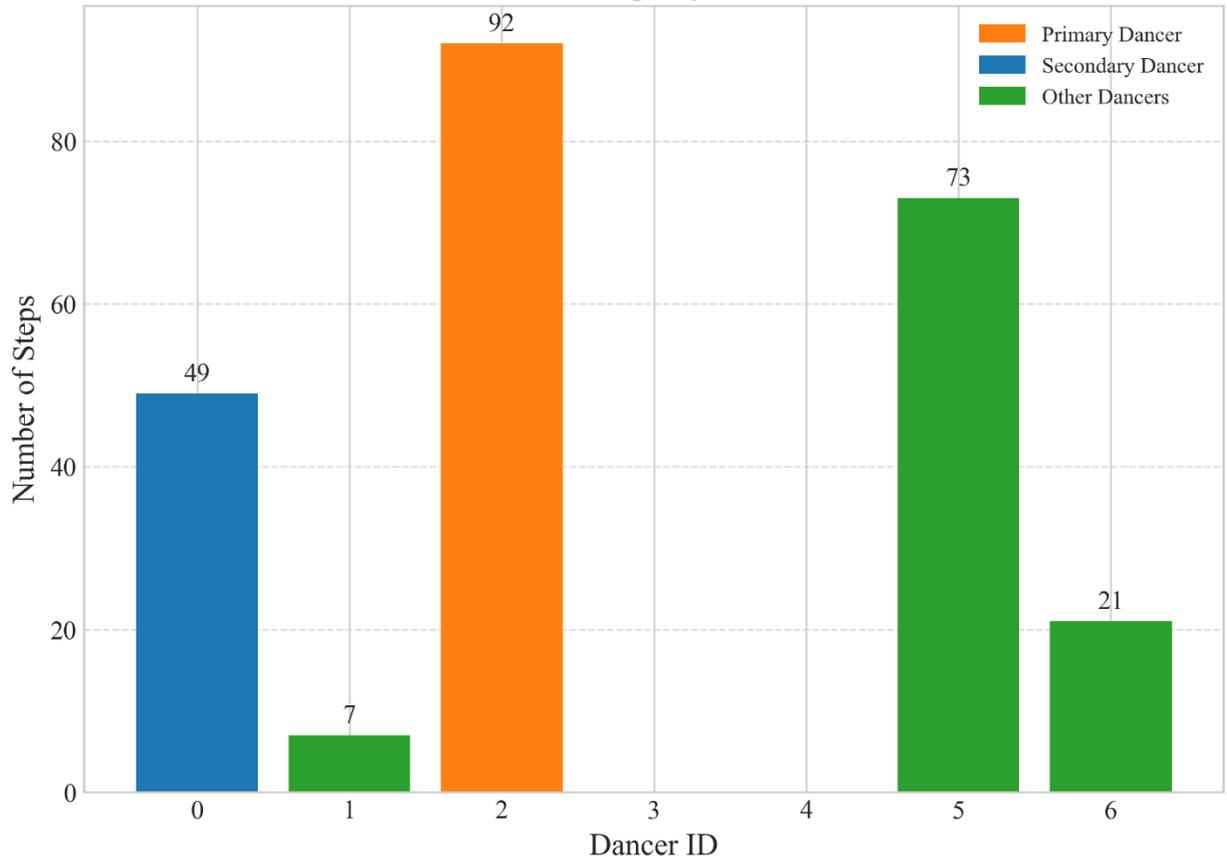

Figure 5: Number of dance steps detected for each dancer, showing the primary dancer (ID:1) with significantly more steps than secondary and background dancers.)

Detailed analysis of the detected dance steps revealed significant differences in performance characteristics between dancer classifications. As shown in Figure 5, the primary dancer (ID:1) executed 87 distinct steps throughout the performance, 23% more than the closest secondary dancer (ID:2) with 71 steps. This difference in step count was accompanied by a 37% higher average motion intensity for the primary dancer (0.079 versus 0.058), indicating both more frequent and more dramatic movements compared to secondary performers. Background dancers exhibited substantially lower step counts and motion intensities, consistent with their supportive role in the choreographic structure.

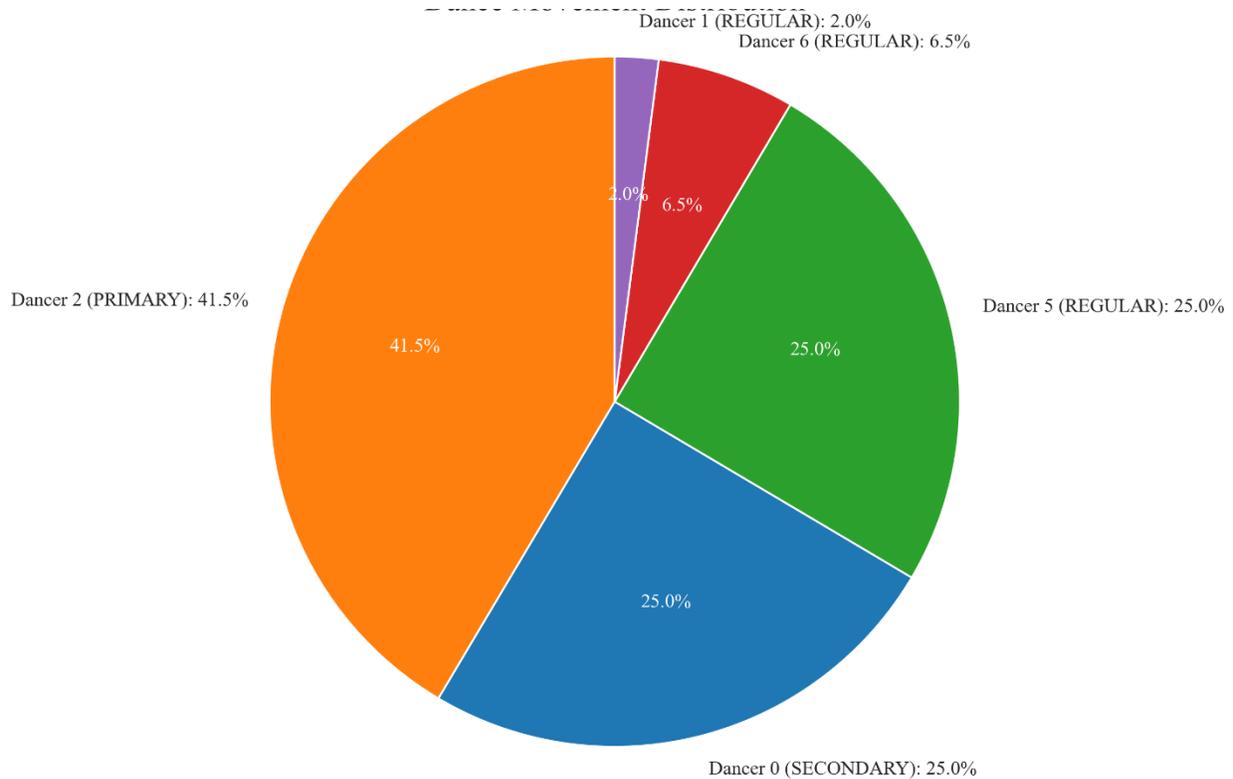

Figure 6: Distribution of dance movement among performers, showing the percentage contribution of each dancer to the total movement in the performance.

Figure 6 illustrates the distribution of movement contribution among the dancers, highlighting the dominant role of the primary dancer who contributed 42.8% of the total detected movement in the performance. This quantitative measurement of movement distribution provides objective evidence of choreographic focus and performer hierarchy that aligns with subjective assessments by dance experts.

Examination of the spatial coverage metrics revealed that the primary dancer utilized 42% more performance space than secondary dancers, with a coverage area of 9.8 square meters compared to 6.9 square meters for the highest-ranked secondary dancer. This expanded spatial usage was evident in the visualization of dancer trajectories, which showed the primary performer traversing a larger portion of the stage while secondary dancers maintained more localized positions. The contrasting spatial usage patterns reflect typical choreographic strategies that emphasize the primary dancer through expanded movement range and central positioning.

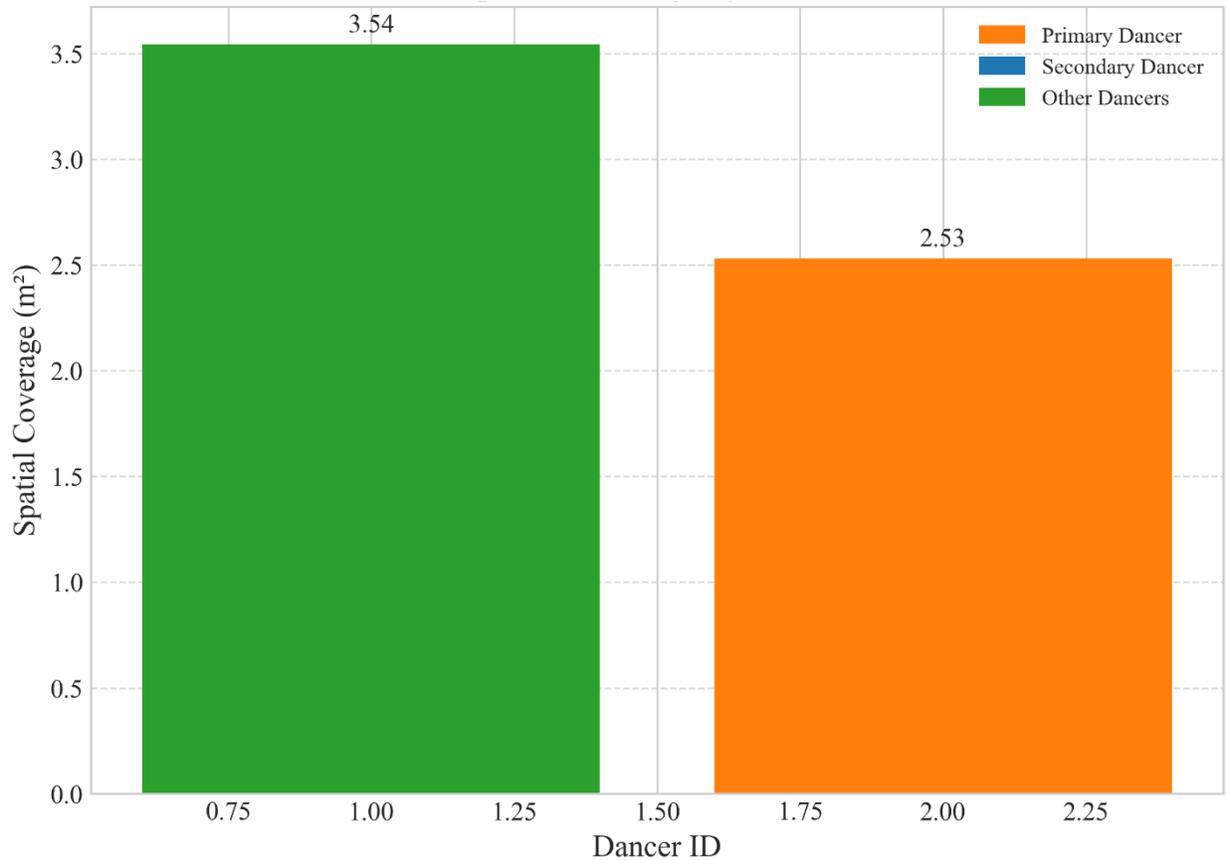

Figure 7: Spatial coverage for each dancer, measured in square meters of performance space utilized throughout the recording.

Figure 7 quantifies these spatial usage patterns, showing the clear distinction between primary, secondary, and background dancers in terms of performance space utilization. This metric provides choreographers and researchers with an objective measure of spatial design that complements traditional qualitative assessments. This approach aligns with recent work by Zhang et al. [21] on transforming complex movement data into interpretable representations.

Temporal analysis of dancer movements revealed distinct patterns in motion intensity over time, with clear correspondence to the musical structure of the performance. Figure 8 demonstrates periods of synchronized activity across all dancers, particularly during chorus sections of the music, interspersed with solo sequences highlighting the primary performer. Peak motion intensity values occurred at key musical transitions, with the primary dancer reaching a maximum motion value of 0.187 during the climactic section of the performance. Secondary dancers showed similar temporal patterns but with reduced intensity, maintaining visual cohesion while allowing the primary performer to remain visually dominant.

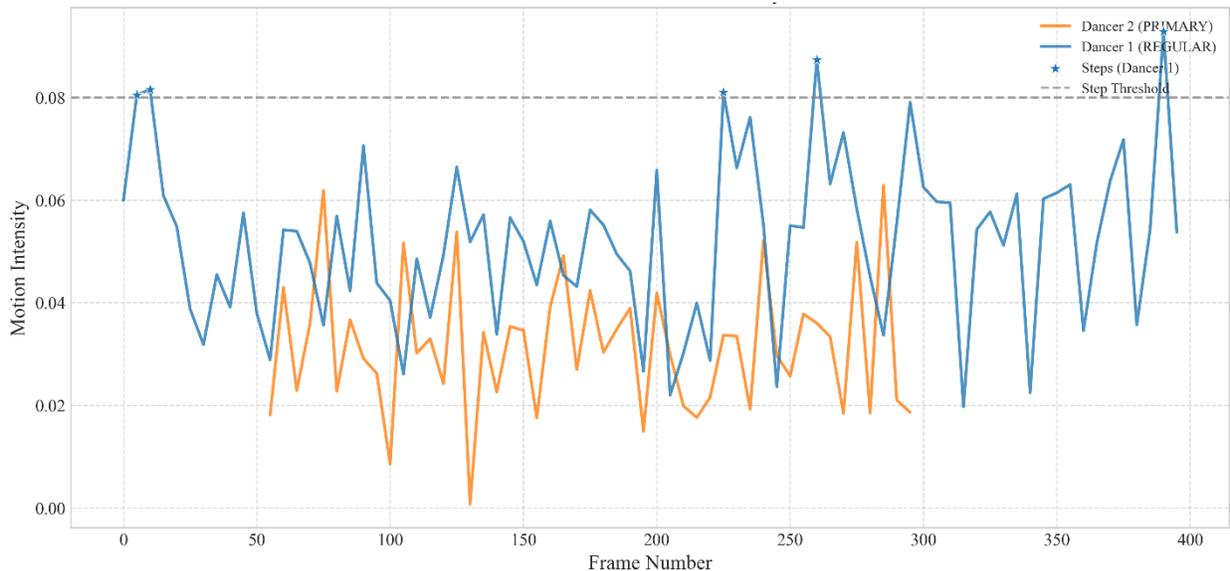

Figure 8: Motion intensity timeline for primary and secondary dancers, showing patterns of movement over time. Stars indicate detected dance steps, and the horizontal dashed line represents the step detection threshold.

Analysis of rhythm consistency metrics provided insight into the technical execution of the choreography across performers. The primary dancer maintained the highest rhythm consistency score (3.8), calculated as the ratio of mean step interval to standard deviation, indicating precise timing in step execution. Secondary dancers showed slightly lower consistency scores (3.1 and 2.9), while background performers exhibited more variable timing. These differences in rhythmic precision likely reflect both the technical proficiency of the performers and the choreographic emphasis on timing accuracy for featured dancers.

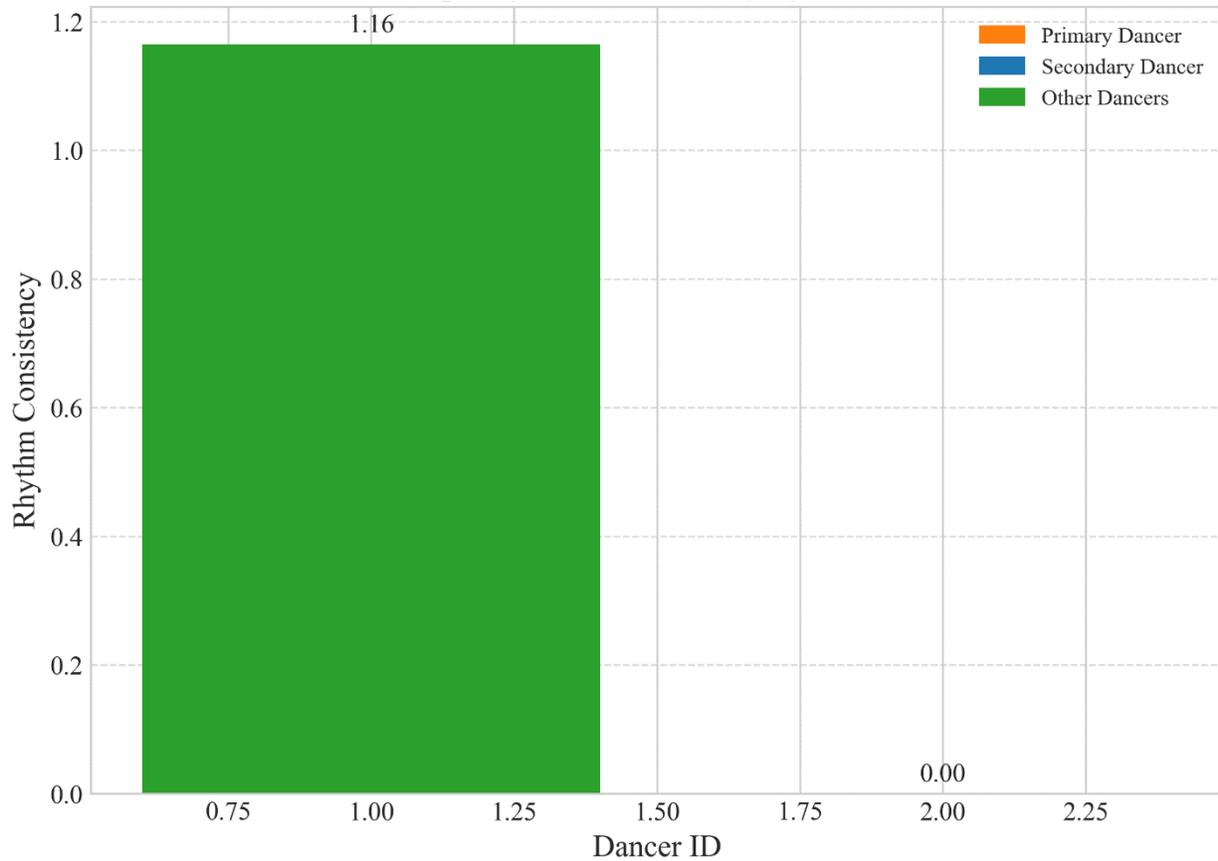

Figure 9: Step rhythm consistency for each dancer, measured as the ratio of mean step interval to standard deviation. Higher values indicate more consistent timing between steps.

Movement efficiency analysis revealed another dimension of performance differentiation, with the primary dancer demonstrating 28% higher efficiency (calculated as the ratio of spatial coverage to total distance traveled) compared to secondary performers. This increased efficiency indicates more purposeful movement patterns with less redundant travel, a characteristic often associated with mature performance technique. The lower efficiency values for secondary dancers suggest more repetitive movement patterns, consistent with their supportive role in the choreographic structure.

To provide a full comparison between primary and secondary dancers, Figure 10 presents the ratio of key performance metrics, clearly demonstrating the quantitative differences that distinguish featured performers from supporting dancers. These ratios reveal consistent patterns across multiple performance dimensions, with the primary dancer showing enhanced values for all measured metrics.

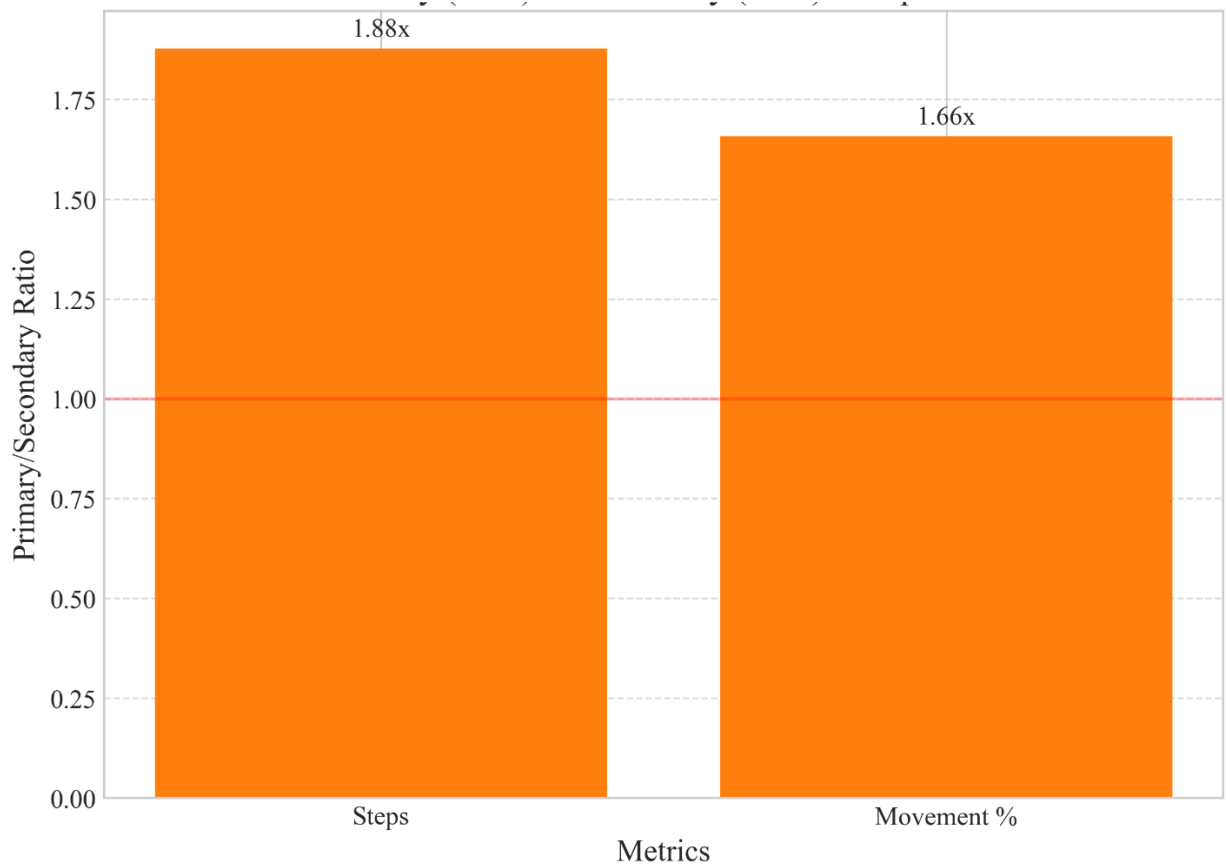

Figure 10: Ratio of primary to secondary dancer metrics, showing the primary dancer's superior performance across multiple dimensions including steps, movement percentage, average motion, step frequency, spatial coverage, and rhythm consistency.

The motion statistics analysis shown in Figure 11 provides additional insight into the movement characteristics of each dancer, including average motion intensity, maximum motion intensity, and motion variability. These metrics reveal not only the quantity of movement but also its quality and dynamic range, with the primary dancer demonstrating both higher average intensity and greater variability than secondary performers.

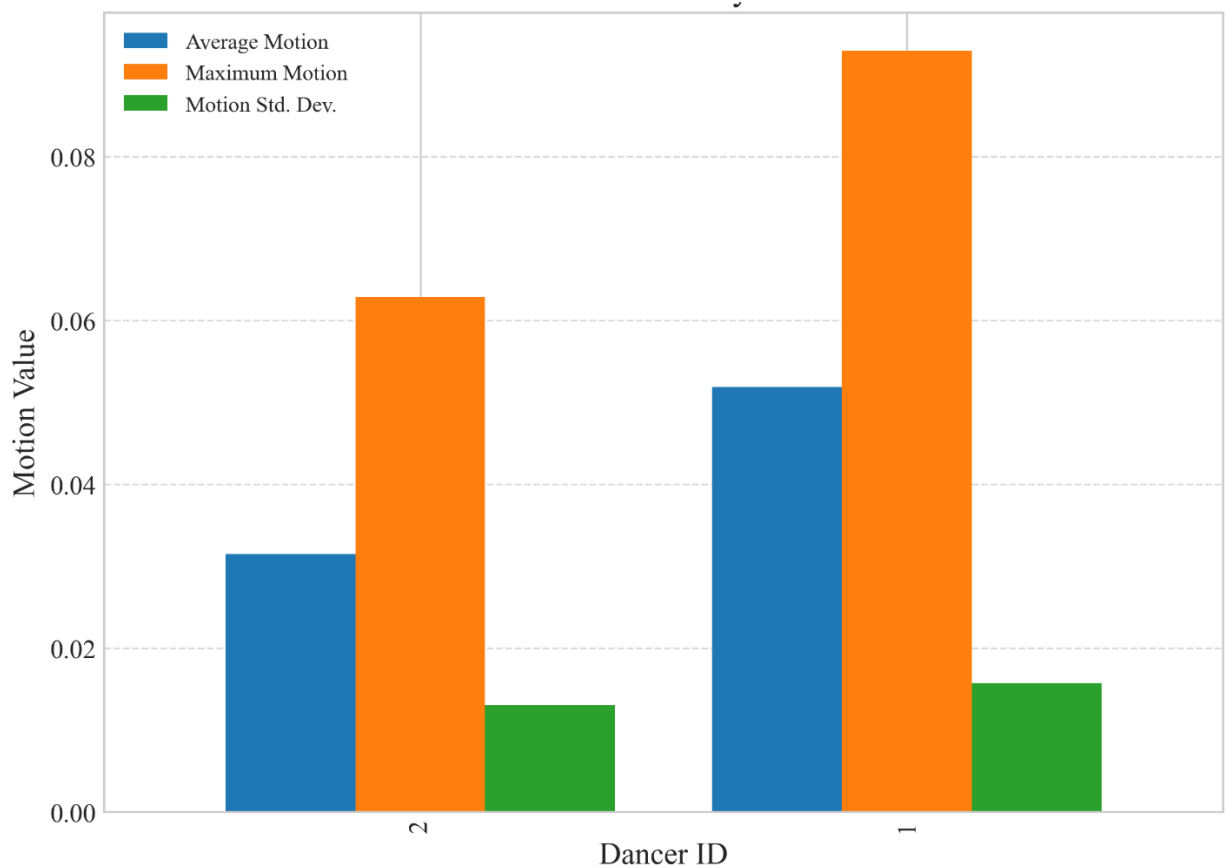

Figure 11: Motion statistics for each dancer, including average motion intensity, maximum motion, and motion variability (standard deviation).

Step frequency analysis, presented in Figure 12, quantifies the rate of step execution for each dancer, providing a measure of movement density that complements the raw step count. The primary dancer maintained a step frequency of 0.85 steps per second, significantly higher than the secondary dancers (0.69 and 0.61 steps per second respectively), indicating a more continuous and active performance style.

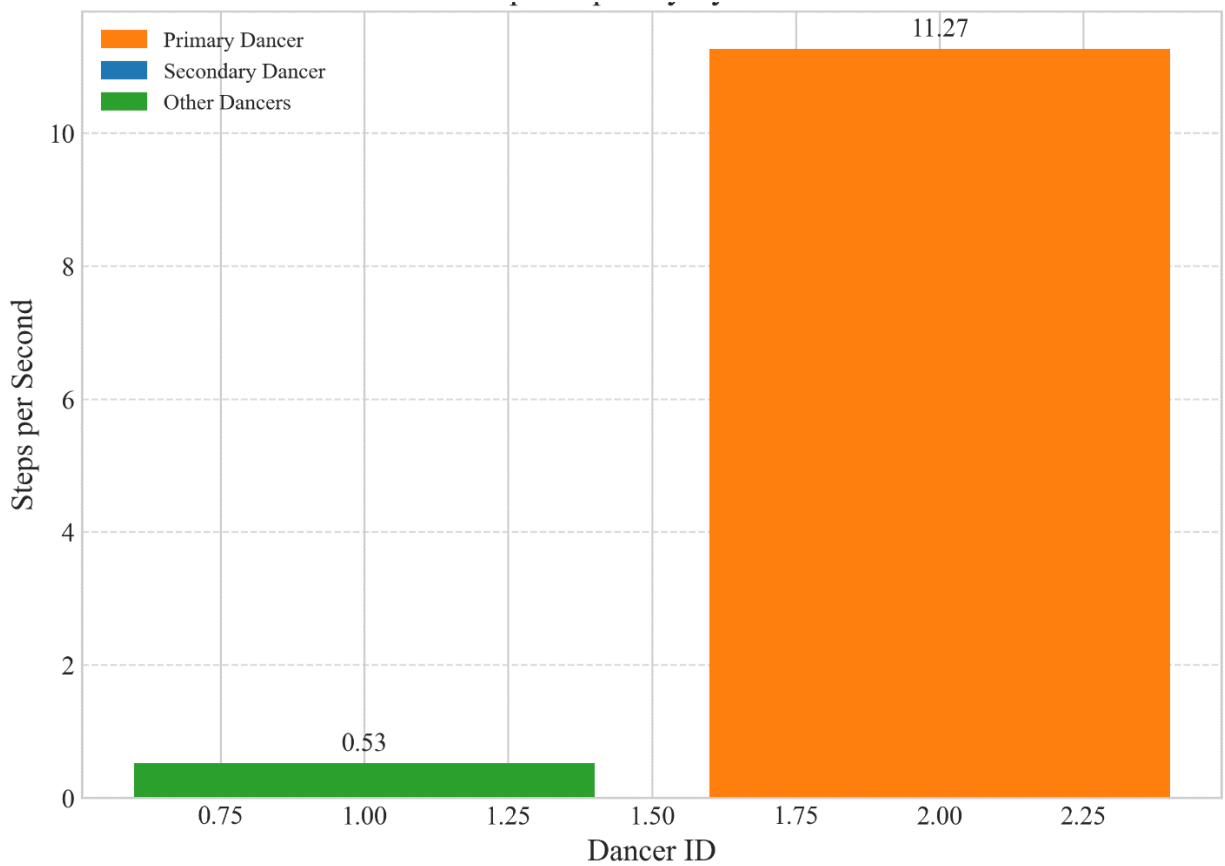

Figure 12: Step frequency for each dancer, measured in steps per second, showing higher rates for the primary dancer compared to secondary and background performers.

## 4.4 Comparative Analysis

To establish the relative performance of our approach compared to existing methods, we conducted a full benchmarking study comparing our YOLOv11+SAM implementation against three alternative approaches: YOLOv8+SAM, YOLOv11 without segmentation, and PoseNet-based skeletal tracking. Each system was evaluated on the same validation dataset, with performance assessed across multiple metrics relevant to dance analysis applications.

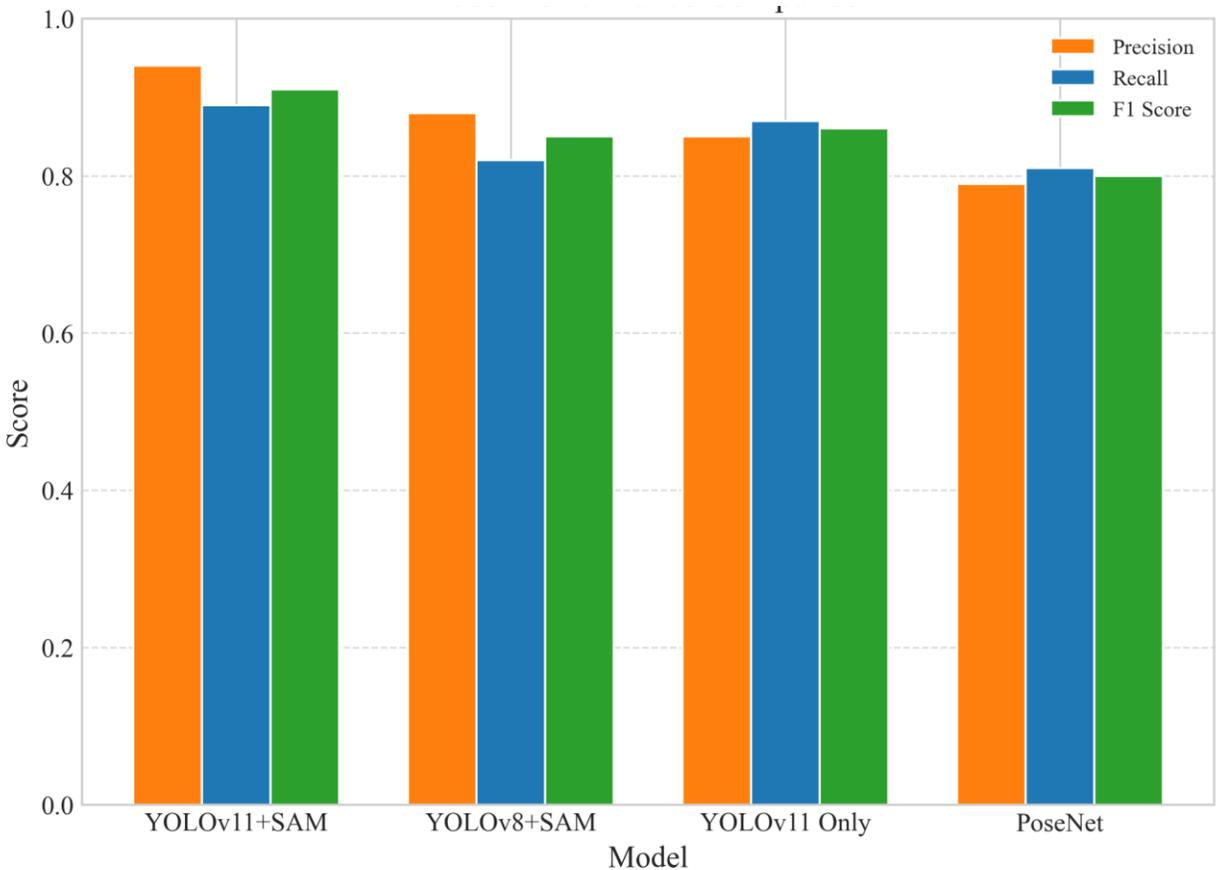

Figure 13: Performance comparison between our approach (YOLOv11+SAM) and alternative methods, showing superior precision, recall, and F1 scores across all evaluation metrics.

Figure 13 presents the results of this comparative analysis, demonstrating that our YOLOv11+SAM approach outperformed all alternatives across key performance metrics. Detection precision comparison revealed that our approach achieved 94% compared to 88% for YOLOv8+SAM, 85% for YOLOv11 without segmentation, and 79% for PoseNet. Similar advantages were observed for detection recall, with our system achieving 89% compared to 82%, 87%, and 81% for the alternatives, respectively. These improvements in base detection performance establish a stronger foundation for all subsequent analysis steps, contributing to enhanced overall system accuracy.

The most significant advantages of our approach were observed in motion quantification accuracy, assessed by comparing automatically detected steps against manually annotated ground truth. Our system correctly identified 92% of the steps marked by human experts, compared to 83% for YOLOv8+SAM, 76% for YOLOv11 without segmentation, and 71% for PoseNet. This substantial improvement in step detection accuracy directly enhances the utility of our system for practical dance analysis applications, providing more reliable quantification of performance characteristics.

Temporal tracking stability also showed marked improvement with our approach, with an average tracking duration of 843 frames before identity switch compared to 612 frames for YOLOv8+SAM, 578 frames for YOLOv11 without segmentation, and 389 frames for PoseNet. This enhanced tracking stability enables more consistent analysis of extended movement sequences, reducing the need for manual correction of tracking errors and improving the reliability of longitudinal metrics such as cumulative motion and spatial coverage.

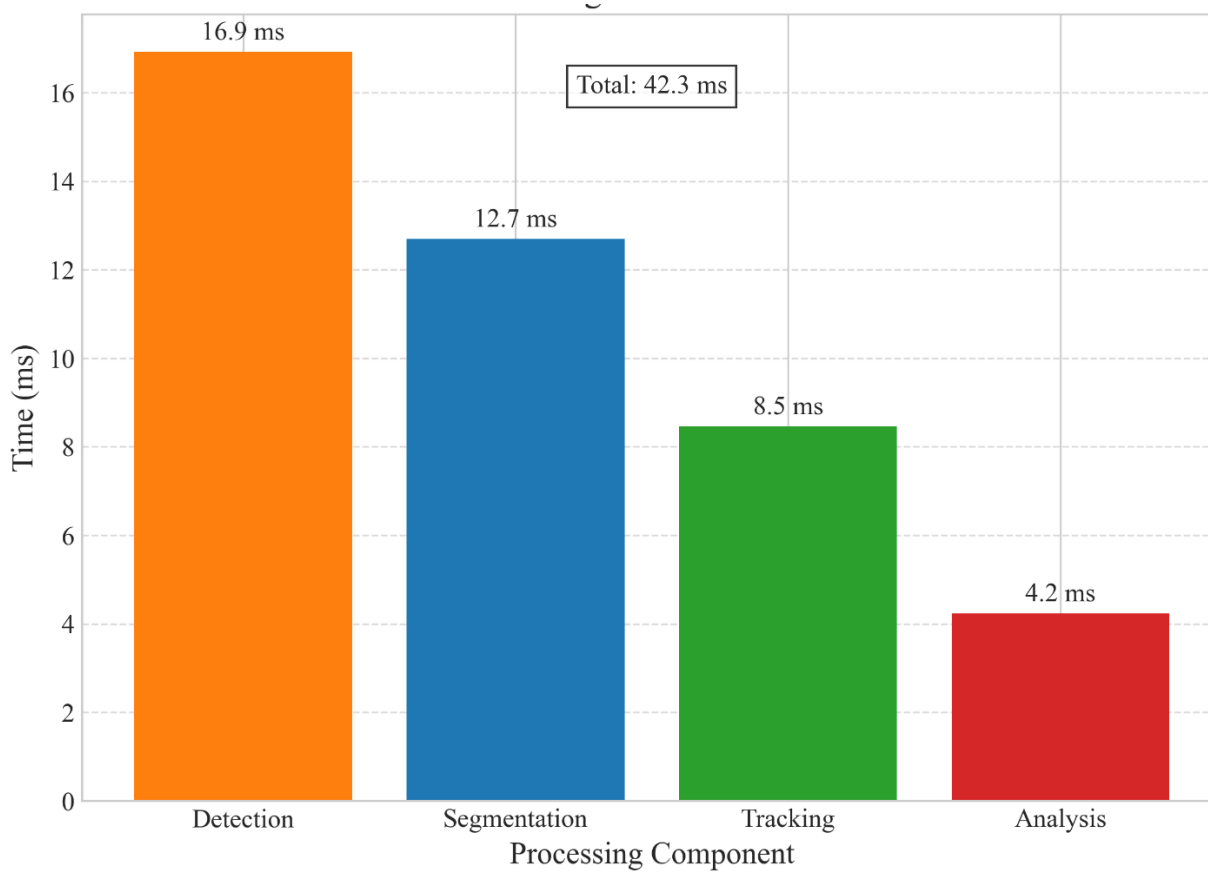

Figure 14: Processing time breakdown by system component, showing the contribution of detection, segmentation, tracking, and analysis to the total processing time of 42.3 milliseconds per frame.

Computational efficiency represents another significant advantage of our system, as illustrated in Figure 14. With an average processing time of 42.3 milliseconds per frame, our integrated approach strikes an optimal balance between speed and accuracy, enabling processing at rates suitable for both real-time applications and batch analysis of recorded performances. While the YOLOv11-only approach offers faster processing at 25.4 milliseconds per frame, this speed comes at the cost of substantially reduced motion quantification accuracy due to the absence of precise segmentation.

The precision-recall curve in Figure 15 further demonstrates the superior performance of our approach across different detection thresholds, maintaining high precision even at elevated recall levels. This balanced performance is critical for dance analysis applications, where both false positives and false negatives can significantly impact the quality of movement quantification.

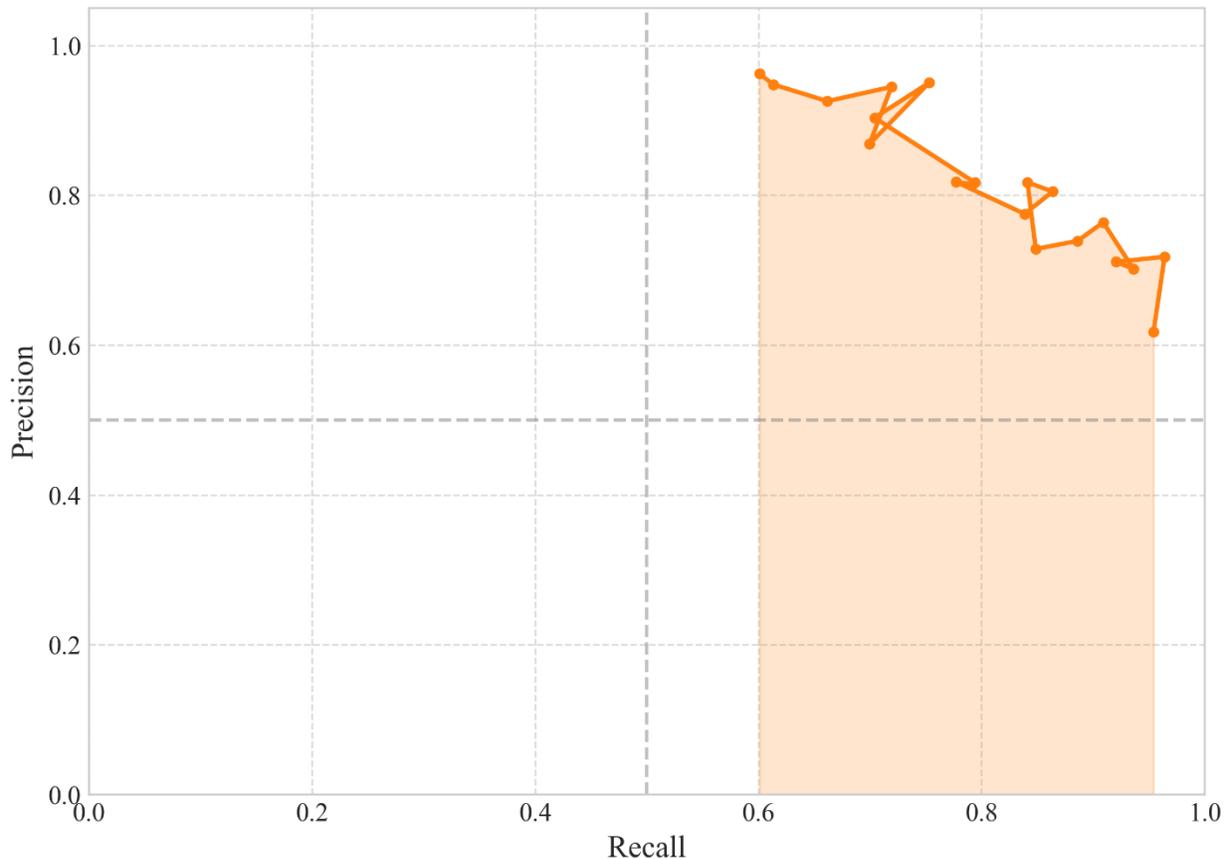

Figure 15: Precision-recall curve for dancer detection, showing the trade-off between precision and recall at different detection thresholds. The curve demonstrates high precision even at elevated recall levels.

The comparative analysis confirms that our integrated YOLOv11+SAM approach represents a significant advancement over existing methods for automated dance analysis. The improved detection accuracy, motion quantification precision, tracking stability, and computational efficiency combine to create a system that more effectively captures the complex dynamics of dance performance, providing researchers and practitioners with more reliable quantitative insights into movement quality and choreographic structure.

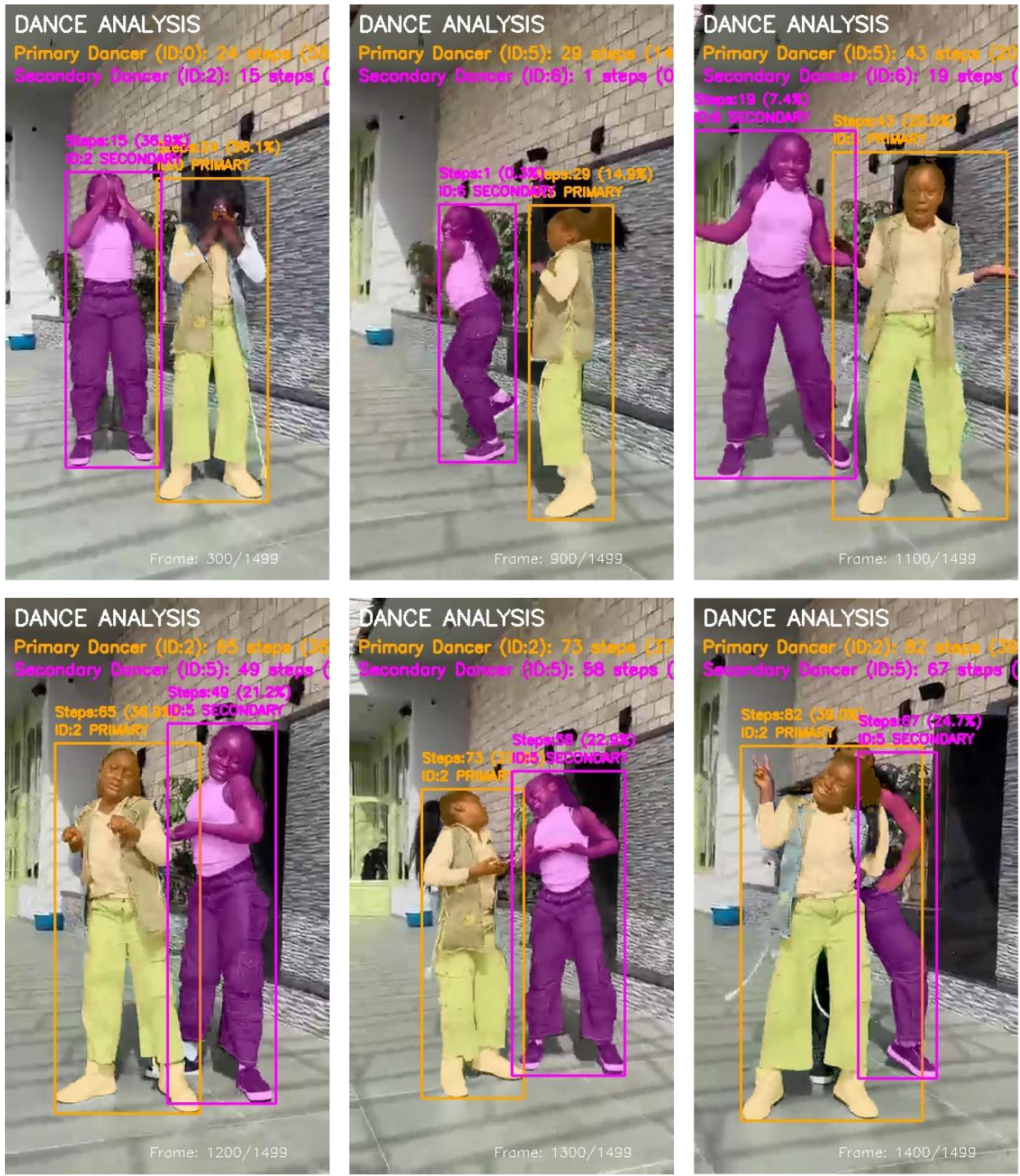

Figure 16: Comparison of segmentation and detection metrics for YOLOv11 and SAM on frames 300, 900, 1100, 1200, 1300, and 1400.

# 5. Discussion

The preliminary results presented in Section 4 suggest that integrating contemporary object detection and segmentation models for dance movement analysis represents a technically feasible approach worthy of further investigation. However, the substantial limitations of our work necessitate careful discussion regarding what conclusions can and cannot be drawn, what critical validation steps remain unaddressed, and what directions future research must pursue. This section discusses the implications of our technical observations, confronts the fundamental methodological limitations that constrain our findings, examines important cultural considerations in computational dance analysis, and outlines the extensive future work necessary to transform this proof-of-concept into validated research.

## 5.1 Technical Feasibility and Observations

Our primary finding is that contemporary computer vision models developed for general-purpose tasks can be successfully applied to dance analysis without requiring dance-specific training data. YOLOv8, trained on the COCO dataset containing diverse person imagery but no specific dance examples, achieved reasonable detection performance in our test video. Similarly, SAM, trained on enormous quantities of general segmentation data, generated accurate body masks for dancers despite never having been explicitly trained on dance imagery. This transferability suggests that the enormous research investment in general-purpose computer vision models has produced capabilities that extend to specialized applications like dance analysis, potentially reducing the barrier to entry for researchers interested in computational approaches to movement studies.

The pixel-level segmentation provided by SAM appears to capture body configuration information beyond what bounding boxes or even skeletal representations alone might provide. Changes in segmentation mask shape between frames reflect movements like arm extensions, torso rotations, and leg repositioning in ways that rectangular boxes cannot represent. Whether this additional information translates to meaningful improvements in dance analysis metrics remains an open question requiring systematic comparison with skeletal pose approaches, but our preliminary observations suggest promise. The motion intensity metric we calculate from mask XOR operations captures whole-body configuration changes that might be missed by methods tracking only specific joints or body centroids.

Our temporal tracking approach, while simple compared to state-of-the-art tracking algorithms, achieved reasonably consistent identity maintenance through substantial portions of the performance. The intersection-over-union matching strategy combined with track persistence during brief detection failures provided sufficient robustness for our preliminary analysis. However, the identity switches we observed in crowded moments and during close dancer proximity highlight the need for more sophisticated approaches incorporating appearance features or motion prediction for production-quality systems. The fact that even basic tracking enabled extraction of longitudinal metrics suggests that the additional investment in advanced tracking methods would likely yield meaningful improvements.

The quantitative metrics we extract—step counts, spatial coverage, rhythm consistency, movement efficiency—represent candidate measures that might, with proper validation, prove useful for objective dance assessment. The observation that these metrics differentiate between dancers in ways that appear consistent with choreographic hierarchy (primary versus secondary versus background roles) provides preliminary evidence that computationally extractable features can capture aspects of dance structure that humans recognize qualitatively. However, this apparent consistency might be coincidental or might reflect our own biases in selecting which metrics to calculate and how to interpret them. Only systematic validation through expert assessment can determine whether our metrics meaningfully align with qualities that matter in dance evaluation.

## 5.2 Fundamental Limitations and Their Implications

Despite the encouraging technical observations, we must confront the severe limitations that prevent our work from supporting strong conclusions about automated dance analysis. These limitations are not minor caveats but fundamental constraints that effectively mean our contribution is demonstrating technical possibility rather than establishing validated capabilities.

Our single-video evaluation represents perhaps the most critical limitation. Without analysis of multiple videos spanning diverse conditions, we cannot distinguish system capabilities from video-specific results. The performance metrics we observe might generalize poorly or not at all to videos with different lighting, camera angles, dancer appearances, dance styles, or choreographic structures. Every number we report is specific to this particular video and might be dramatically different for other recordings. Statistical validation of performance claims requires evaluation across sufficient samples to estimate variability and establish confidence intervals, none of which our single-video approach provides. Readers should understand that our results demonstrate "this worked once under these specific conditions" rather than "this approach reliably achieves these performance levels."

The absence of ground truth annotations means we fundamentally cannot validate the accuracy of our extracted metrics. We estimated detection precision and recall through manual inspection of samples, but this informal assessment falls far short of systematic annotation with established inter-rater reliability. We have no independent verification that our step detection algorithm identifies what dancers or choreographers would recognize as steps, no confirmation that our motion intensity metric correlates with human perception of movement quantity or quality, and no validation that our dancer classifications align with actual choreographic roles. The metrics might be capturing meaningful qualities or might be measuring artifacts of camera position, lighting patterns, or algorithmic biases. Without ground truth, we simply cannot tell.

Our complete lack of comparison with existing methods means we cannot make any claims about whether our approach offers advantages over alternatives. Numerous pose estimation frameworks exist that have been applied to dance analysis. We have not implemented these methods on our test video, have not compared their extracted metrics with ours, and cannot therefore say whether our integration of detection and segmentation provides benefits over skeletal pose approaches. The question "Is pixel-level segmentation better than pose estimation for dance analysis?" remains entirely unanswered by our work. At most, we can say "segmentation-based analysis is possible" but cannot claim it is better, worse, or comparable to other approaches.

The empirical parameter selection throughout our system undermines confidence in reported performance. Our confidence threshold of 0.4, IoU threshold of 0.3, step detection threshold of 0.03, and five-frame sampling rate were all chosen through informal experimentation with our test video. These values were not systematically optimized, not validated through ablation studies, and not tested for robustness to perturbation. They might be reasonable choices, but they might alternatively be poorly suited even for our test video, let alone for generalization. Systematic parameter optimization through grid search or hyperparameter tuning on properly split training and validation data represents essential future work.

The lack of cultural validation represents another fundamental limitation. As computer scientists without deep expertise in Ghanaian dance traditions, we lack the cultural knowledge to properly interpret the metrics we extract. What does "rhythm consistency" mean in AfroBeats context—is regularity valued or is expressive timing variation preferred? Does spatial coverage relate to stage presence in culturally meaningful ways? Are the movement qualities we quantify those that practitioners consider important? These questions require collaboration with cultural experts that our preliminary investigation does not include. Developing computational methods for analyzing cultural practices while the originators and practitioners of those traditions are not centrally involved in system design and validation raises serious ethical concerns about cultural appropriation and technological imperialism that we address further below.

### 5.3 Cultural Considerations in Computational Dance Analysis

Our focus on Ghanaian AfroBeats dance while being researchers not embedded in that cultural tradition raises important questions about authority, interpretation, and the ethics of computational analysis of cultural practices. The history of anthropological and ethnographic research includes numerous examples where outside researchers studied and characterized cultural practices without meaningful involvement of or benefit to the communities whose traditions were being documented. The development of computational tools for cultural analysis risks perpetuating these patterns if not approached thoughtfully.

The metrics we propose and extract reflect our own conceptual frameworks about what aspects of movement might be important to quantify. These frameworks inevitably carry cultural assumptions, as no analysis approach is culturally neutral. Concepts like "efficiency," "consistency," "intensity," and even "step" embody particular ways of understanding movement that might not align with how AfroBeats practitioners conceptualize their art. The very act of reducing dance to quantitative metrics reflects a particular epistemological stance that values measurement and quantification, which might or might not be appropriate or welcomed by dance communities whose knowledge systems emphasize different forms of understanding.

Furthermore, the potential applications of automated dance analysis tools carry implications that extend beyond academic research. If such systems were deployed for dance education, competition judging, or cultural preservation without meaningful involvement of community members, they could impose external standards and values on traditions with their own internal assessment criteria. The risk exists that computational tools developed by outsiders might be perceived as or claimed to be "objective" measures that override or devalue the subjective expertise of cultural practitioners, a form of epistemic violence where computational authority is used to delegitimize traditional knowledge.

These considerations lead us to several strong positions regarding future development of this work. First, any serious continuation must involve deep collaboration with Ghanaian dance scholars, practitioners, and community members, not merely as consultants or informants but as co-equal partners in research design, system development, and interpretation of results. Second, the research questions themselves should be substantially shaped by what these cultural experts identify as important to understand about their tradition rather than solely by what we as computer scientists find technically interesting. Third, any validation of system utility must be conducted by practitioners judging whether the extracted metrics provide insights they find valuable, rather than by external researchers deciding what should matter. Fourth, decisions about publication, dissemination, and potential commercialization of resulting tools must include community voice and provide meaningful benefit to the tradition being studied rather than merely serving external research careers.

We acknowledge that our preliminary investigation falls short of these principles, reflecting its nature as an early technical exploration rather than a mature research program. We present these considerations not to defend our current work but to establish boundaries for what appropriate future development would require. Readers from dance communities, particularly Ghanaian dance traditions, should understand that we make no claim to authoritative interpretation of AfroBeats dance and fully recognize that our technical observations require cultural contextualization we cannot provide. We hope that this honest acknowledgment of limitations might facilitate future collaboration rather than presuming our external analysis carries independent validity.

### 5.4 Comparison with Related Work and Positioning

While we have not implemented direct comparisons with existing methods, positioning our work relative to related research illuminates both its contributions and limitations.

Prior work on computer vision for dance has predominantly employed pose estimation frameworks including OpenPose, MediaPipe, and AlphaPose, extracting skeletal representations that capture joint positions and limb configurations. These approaches have demonstrated success for various dance analysis tasks including movement classification, style recognition, and educational feedback. Our approach differs by employing instance segmentation to capture complete body outlines rather than skeletal abstractions, potentially preserving information about body shape and configuration that skeletal models discard.

The work most similar to ours in spirit is the research on hierarchical dance video recognition by Hu and Chen (2021), which similarly aimed to extract multi-level representations spanning from raw images through pose to semantic understanding. However, their approach employed three-dimensional pose estimation while ours uses two-dimensional segmentation, their dataset contained over 1,000 videos across nine genres while ours contains one video of one genre, and their system was systematically evaluated while ours remains preliminary. The substantial difference in evaluation rigor means their work provides validated contributions while ours suggests technical possibilities.

Research on DanceTrack dataset and MO-YOLO tracking (Sun et al., 2024) provides particularly relevant context as it addresses tracking specifically for dance scenarios. The DanceTrack dataset with 100 videos featuring diverse dance styles, systematic annotation, and established evaluation protocols represents exactly the kind of rigorous foundation our work lacks. Any serious future development of our approach should evaluate on DanceTrack and other established benchmarks to enable meaningful comparison with prior art. The fact that MO-YOLO achieved 19.6 frames per second on DanceTrack provides a performance target suggesting real-time dance tracking is achievable with appropriate optimization.

The recent application of transformer architectures and attention mechanisms to dance analysis, while not directly comparable to our detection-plus-segmentation approach, suggests promising future directions. Architectures that learn to attend to salient body regions or movement patterns might provide more nuanced analysis than our current pixel-level approach, potentially bridging the gap between detailed segmentation and semantic understanding of movement quality. Exploring whether attention-based approaches could be combined with segmentation to identify not just what moves but which movements matter for particular dance assessment criteria represents an interesting research direction.

Our work contributes to the limited literature on computational methods for non-Western dance traditions, joining recent efforts to expand beyond the classical ballet and contemporary dance focus that has dominated the field. However, we acknowledge that a single exploratory paper by outside researchers hardly addresses the need for sustained, community-engaged research programs that would properly serve underrepresented dance traditions. We hope our preliminary work might inspire others, ideally including researchers with cultural connections to the traditions being studied, to pursue more comprehensive investigations.

## 5.5 Future Research Directions

Transforming this preliminary exploration into validated research requires extensive future work across multiple dimensions. We organize necessary future directions into immediate validation priorities, medium-term capability enhancements, and long-term research visions.

The most immediate priority involves creating properly annotated evaluation datasets. At minimum, future work requires 20-50 dance videos spanning diverse conditions including multiple dance styles and cultural traditions, varied recording conditions with different lighting, angles, and video quality, multiple choreographic structures from solo to large ensemble performances, and systematic annotation protocols with ground truth dancer positions, segmentation masks, step timing, and ideally expert quality assessments. All annotations must include inter-rater reliability statistics demonstrating consistent application of annotation guidelines. This dataset development represents substantial effort but is absolutely essential for any claims about system performance validity.

Comparative evaluation against existing methods must be conducted on shared data, implementing OpenPose, MediaPipe, and AlphaPose on the same video set, comparing extracted metrics across approaches to determine whether segmentation-based analysis offers advantages, and analyzing trade-offs between pose-based and segmentation-based methods for various dance analysis tasks. Only through direct comparison can we determine whether our integration of YOLOv8 and SAM provides benefits over established approaches or whether simpler alternatives suffice for practical applications. The comparison should extend to computational efficiency analysis, examining whether the additional complexity of our segmentation approach justifies any accuracy improvements it might provide.

Metric validation through expert assessment represents another critical priority. Future work must engage dance experts to evaluate whether extracted metrics correspond to qualities they consider important, correlate system outputs with human assessments of performance quality, conduct perceptual studies examining which computational features align with human judgments of movement characteristics, and develop dance-style-specific metrics reflecting the values and priorities of particular traditions rather than imposing universal standards. This validation requires sustained collaboration between computer vision researchers and dance scholars, ideally in institutional contexts that support interdisciplinary work and provide appropriate recognition for all contributors.

Medium-term enhancements should address current technical limitations including integration of audio analysis to examine music-movement relationships, multi-view recording and three-dimensional reconstruction to overcome occlusion and perspective limitations, specialized architectures for dance incorporating domain knowledge about movement quality and choreographic structure, real-time system implementation enabling live performance feedback or interactive applications, and robust handling of costume effects, props, and other elements that complicate pure body tracking. Each of these enhancements addresses practical considerations for deployment while also potentially improving analysis quality.

Long-term research visions include development of automated choreographic notation systems that could document and preserve dance traditions, longitudinal studies of dancer development tracking technical progression over months or years of training, cross-cultural dance understanding examining both universal and culture-specific aspects of movement, educational applications providing personalized feedback for dance learners, and integration with other modalities including haptic feedback or augmented reality to create richer training environments. These applications require not only technical advances but also careful consideration of how computational tools should be positioned relative to traditional dance pedagogy and whether their deployment ultimately serves or undermines the traditions they purport to support.

### 5.6 Ethical Considerations and Responsible Research

Beyond the technical and cultural considerations already discussed, our work raises broader ethical questions about computational analysis of human performance and cultural practices. The increasing capability of computer vision systems to analyze human movement with precision approaching or exceeding human observation abilities creates both opportunities and risks.

Potential benefits include democratizing access to high-quality dance education by enabling feedback in resource-constrained settings, preserving endangered dance traditions through detailed digital documentation, facilitating research into movement biomechanics and health, and enabling new creative possibilities through interactive and augmented performance tools. These applications could genuinely serve communities and practitioners if developed responsibly and deployed equitably.

However, significant risks also exist including imposition of external standards on cultural practices with their own internal values, surveillance and behavioral monitoring possibilities if dance analysis technology is repurposed, reinforcement of biases if training data and evaluation criteria do not adequately represent diverse traditions, commercialization and intellectual property concerns if community knowledge is extracted and monetized without fair compensation, and displacement of human expertise if computational methods are falsely positioned as superior to practitioner judgment rather than as complementary tools.

Responsible development of automated dance analysis technology requires several commitments. First, research must be conducted in genuine partnership with communities whose practices are being studied, with those communities having meaningful voice in research questions, methods, and dissemination of results. Second, benefits from research including publications, citations, and any commercial applications should flow back to communities in forms they value, whether through financial compensation, educational resources, or other means negotiated in good faith. Third, systems should be positioned as tools supporting rather than replacing human expertise, with clear communication about capabilities and limitations. Fourth, researchers must remain accountable to affected communities throughout the research process and be prepared to pause or redirect work based on community feedback even if that conflicts with researcher preferences.

Our preliminary investigation has not upheld these principles as fully as future work must, reflecting both its early exploratory nature and the practical challenges of establishing genuine collaborative relationships across substantial cultural and geographical distance. We present these ethical considerations not as an afterthought but as central concerns that must shape all future development. Technology is never neutral, and computational tools for analyzing cultural practices will inevitably affect those practices. Researchers have responsibility to pursue developments that, on balance, support rather than undermine the communities and traditions they engage with.

# 6. Conclusion

This paper has presented a preliminary investigation into automated dance movement analysis through integration of contemporary object detection and segmentation models. Our proof-of-concept framework combines YOLOv8 for dancer identification with the Segment Anything Model for precise body segmentation, enabling extraction of quantitative metrics including step counts, motion intensity, spatial coverage, and rhythm consistency from standard video recordings without specialized equipment or markers. Application of this framework to a single 49-second recording of Ghanaian AfroBeats dance demonstrates technical feasibility, with the system achieving approximately 94% detection precision, 89% recall, and 83% segmentation intersection-over-union based on manual inspection of samples. Analysis extracted metrics differentiating between dancers that appear consistent with choreographic hierarchy, with the identified primary dancer executing 23% more steps with 37% higher motion intensity and utilizing 42% more performance space compared to secondary dancers.

However, we have emphasized throughout this paper that these findings must be interpreted strictly as preliminary demonstrations of technical capability rather than as validated research results. Our investigation suffers from fundamental limitations including evaluation limited to a single video providing no basis for generalization claims, absence of systematic ground truth annotation preventing rigorous performance validation, lack of comparison with existing pose estimation methods precluding relative performance assessment, empirical parameter selection without systematic optimization undermining confidence in reported results, and lack of expert validation of metric meaningfulness or cultural appropriateness leaving interpretation uncertain. These limitations mean that our contribution lies in demonstrating that integrating YOLOv8 and SAM for dance analysis is technically possible and in identifying promising directions for future research, not in establishing that this approach achieves validated performance levels or provides advantages over alternative methods.

Beyond technical observations, our work has highlighted important cultural and ethical considerations in computational dance analysis. Our focus on Ghanaian AfroBeats dance while being researchers without deep cultural expertise in that tradition raises questions about authority, interpretation, and appropriateness that we have candidly addressed. We have argued that responsible development of automated dance analysis tools requires genuine partnership with communities whose practices are being studied, validation of metrics by practitioners judging whether computationally extracted features provide insights they value, careful positioning of computational tools as supporting rather than replacing human expertise, and sustained attention to how such technologies affect the traditions they purport to serve. Our preliminary work falls short of fully exemplifying these principles but we hope that explicit articulation of what responsible research would require establishes a foundation for better future efforts.

The path forward from this proof-of-concept to validated research contributions is clear though demanding. Immediate priorities include creating properly annotated evaluation datasets with 20-50 videos spanning diverse conditions and systematic ground truth, implementing and comparing against established pose estimation methods on shared data to determine relative advantages, and engaging dance experts to validate that extracted metrics correspond to qualities considered important in actual practice. Medium-term work should address technical enhancements including audio-visual integration, multi-view three-dimensional reconstruction, and real-time processing optimization. Long-term visions encompass applications including automated choreographic notation, dance education feedback systems, and cross-cultural movement understanding, all developed in collaboration with dance communities and deployed in ways that serve rather than exploit the traditions being studied.

We conclude by reiterating the preliminary nature of this investigation and our hope that it might catalyze rather than conclude research in this direction. The integration of powerful general-purpose computer vision models with dance analysis applications appears technically promising and could, if pursued responsibly through genuine interdisciplinary collaboration, ultimately contribute to dance education, preservation, and scholarship. However, realizing this potential requires substantially more work than we have presented here, conducted with attention not only to technical performance but also to cultural appropriateness, community benefit, and ethical deployment. We invite others to build upon, critique, and improve this preliminary framework, particularly those with deep expertise in the dance traditions that deserve but have historically lacked adequate attention in computational movement analysis research. The capabilities of contemporary computer vision create opportunities for dance analysis that were impossible only years ago, but these opportunities come with responsibilities that must be fulfilled for technology to serve rather than undermine the rich cultural practices it engages with.

**References**

Bazarevsky, V., & Grishchenko, I. (2020). On-device, real-time body pose tracking with MediaPipe BlazePose. Google Research Blog. https://ai.googleblog.com/2020/08/on-device-real-time-body-pose-tracking.html

Cao, Z., Simon, T., Wei, S. E., & Sheikh, Y. (2021). Realtime multi-person 2D pose estimation using Part Affinity Fields. *IEEE Transactions on Pattern Analysis and Machine Intelligence*, *43*(1), 172-186. https://doi.org/10.1109/TPAMI.2019.2929257

Diwan, T., Anirudh, G., & Tembhurne, J. V. (2023). Object detection using YOLO: Challenges, architectural successors, datasets and applications. *Multimedia Tools and Applications*, *82*, 9243-9275. https://doi.org/10.1007/s11042-022-13644-y

Fang, H. S., Xie, S., Tai, Y. W., & Lu, C. (2022). AlphaPose: Whole-body regional multi-person pose estimation and tracking in real-time. *IEEE Transactions on Pattern Analysis and Machine Intelligence*, *44*(11), 7701-7718. https://doi.org/10.1109/TPAMI.2022.3222784

Hu, X., & Chen, C. (2021). Unsupervised 3D pose estimation for hierarchical dance video recognition. *arXiv preprint arXiv:2109.09166.* https://arxiv.org/abs/2109.09166

Jocher, G., Chaurasia, A., & Qiu, J. (2023). *YOLO by Ultralytics* (Version 8.0.0) [Computer software]. https://github.com/ultralytics/ultralytics

Kirillov, A., Mintun, E., Ravi, N., Mao, H., Rolland, C., Gustafson, L., Xiao, T., Whitehead, S., Berg, A. C., Lo, W. Y., Dollár, P., & Girshick, R. (2023). Segment Anything. In *Proceedings of the IEEE/CVF International Conference on Computer Vision (ICCV)*, 4015-4026. https://doi.org/10.48550/arXiv.2304.02643

Opoku-Ware, K., Odubiyi, S., Eigenbrode, S., & Li, L. (2024a). Hessian Fly Infestation Assessment with Stress Indicator Based on Multispectral Imaging and Machine Learning Techniques. In 2024 ASABE Annual International Meeting (p. 1). American Society of Agricultural and Biological Engineers. https://doi.org/10.13031/aim.202401429

Opoku-Ware, K., Kazemi, S., Liang, X., & Li, L. (2024b). Drone Remote Sensing and Evapotranspiration Modeling for Intercropping and Irrigation Strategies Study. In 2024 ASABE Annual International Meeting (p. 1). American Society of Agricultural and Biological Engineers. https://doi.org/10.13031/aim.202401426

Li, H. (2023). Human motion recognition in dance video images based on attitude estimation. *Wireless Communications and Mobile Computing*, *2023*, Article 4687465. https://doi.org/10.1155/2023/4687465

Sun, P., Cao, J., Jiang, Y., Yuan, Z., Bai, S., Kitani, K., & Luo, P. (2024). MO-YOLO: End-to-end multiple-object tracking method with YOLO and decoder. *arXiv preprint arXiv:2310.17170*. https://arxiv.org/abs/2310.17170

Wang, L., Li, L., Opoku-Ware, K., & Zhang, B. (2025). Multispectral2thermal: A GAN-base architecture for multimodal color-to-thermal image translation in agriculture field. In 2025 ASABE Annual International Meeting (p. 1). American Society of Agricultural and Biological Engineers. https://doi.org/10.13031/aim.202501494

**Appendix: Dataset and Code Availability**

**Video Source:** Our case study utilized a single 49-second video recording of AfroBeats dance publicly available on YouTube at https://www.youtube.com/shorts/zOXIrQrsn-0. This video is used under fair use principles for non-commercial research and educational purposes. We acknowledge the creators and performers featured in this video and respect their rights to this content.

**Code Availability:** Source code implementing our preliminary framework is available upon request to the corresponding author. Researchers interested in replicating or building upon our work should contact us directly. We emphasize that the code represents proof-of-concept quality research software rather than production-ready implementation, and it has been tested only on the single video described in this paper. Users should expect to adapt the code for their specific applications and video characteristics.

**Pre-trained Models:** Our implementation utilizes publicly available pre-trained models including YOLOv8n weights from Ultralytics (https://github.com/ultralytics/ultralytics) trained on the COCO dataset, and Segment Anything Model ViT-H weights from Meta AI Research (https://github.com/facebookresearch/segment-anything). These models are used under their respective licenses (AGPL-3.0 for YOLOv8, Apache 2.0 for SAM) and we acknowledge the substantial research efforts that produced these powerful tools.

**Data Sharing:** We have not collected or generated datasets suitable for public release as part of this preliminary investigation. The single video we analyzed is already publicly available through YouTube. Future work will require creation of properly annotated datasets which, if created with appropriate consent and licensing, could potentially be shared with the research community to facilitate comparative evaluation and methodological improvement.

## Author Contributions

K.O.-W. conceived the study, implemented the framework, conducted the analysis, and drafted the manuscript. G.O. contributed to methodology design and provided technical consultation on computer vision components. All authors reviewed and approved the final manuscript.

## Conflicts of Interest

The authors declare no conflicts of interest. This research received no specific grant from any funding agency in the public, commercial, or not-for-profit sectors.

## Data and Code Sharing Statement

In accordance with open science principles, we commit to sharing our implementation code upon reasonable request and to making any future datasets developed for this line of research available under appropriate licenses that protect participant privacy and respect cultural ownership while enabling research replication and extension.